\documentclass{article}

\usepackage[preprint]{neurips_2021}

\usepackage[utf8]{inputenc} 
\usepackage[T1]{fontenc}    
\usepackage[hidelinks]{hyperref} 
\usepackage{url}            
\usepackage{booktabs}       
\usepackage{amsfonts}       
\usepackage{amsmath}
\usepackage{amssymb}
\usepackage{graphicx}
\usepackage{caption}
\usepackage{sidecap}
\usepackage{subcaption}
\usepackage{array}
\usepackage{nicefrac}       
\usepackage{microtype}      
\usepackage{xcolor}         
\usepackage{siunitx}
\usepackage{pgfplots}
\pgfplotsset{compat=1.17}

\usepgfplotslibrary{external}
\definecolor{color1}{HTML}{CA0020}
\definecolor{color2}{HTML}{0571B0}

\newtheorem{definition}{Definition}[section]
\newtheorem{theorem}{Theorem}[section]

\DeclareMathOperator{\dive}{div}

\DeclareMathOperator{\diag}{diag}

\DeclareMathOperator{\spn}{span}
\DeclareSIUnit{\nothing}{\relax}

\title{ChebLieNet: Invariant Spectral Graph NNs Turned Equivariant by Riemannian Geometry on Lie Groups}

\author{%
  Hugo Aguettaz \\
  EPFL, Lausanne, Switzerland \\
  \texttt{hugo.aguettaz@epfl.ch} \\
  \AND
  Erik J Bekkers \\
  UvA, Amsterdam, Netherlands \\
  \texttt{e.j.bekkers@uva.nl} \\
  \And
  Michaël Defferrard \\
  EPFL, Lausanne, Switzerland \\
  \texttt{michael.defferrard@epfl.ch} \\
}

\begin{document}

\maketitle

\begin{abstract}
    We introduce ChebLieNet, a group-equivariant method on (anisotropic) manifolds. Surfing on the success of graph- and group-based neural networks, we take advantage of the recent developments in the geometric deep learning field to derive a new approach to exploit any anisotropies in data. Via discrete approximations of Lie groups, we develop a graph neural network made of anisotropic convolutional layers (Chebyshev convolutions), spatial pooling and unpooling layers, and global pooling layers. Group equivariance is achieved via equivariant and invariant operators on graphs with anisotropic left-invariant Riemannian distance-based affinities encoded on the edges. Thanks to its simple form, the Riemannian metric can model any anisotropies, both in the spatial and orientation domains. This control on anisotropies of the Riemannian metrics allows to balance equivariance (anisotropic metric) against invariance (isotropic metric) of the graph convolution layers. Hence we open the doors to a better understanding of anisotropic properties. Furthermore, we empirically prove the existence of (data-dependent) sweet spots for anisotropic parameters on CIFAR10. This crucial result is evidence of the benefice we could get by exploiting anisotropic properties in data. We also evaluate the scalability of this approach on STL10 (image data) and ClimateNet (spherical data), showing its remarkable adaptability to diverse tasks.
\end{abstract}

\section{Introduction} \label{sec:introduction}

Deep learning is a class of machine learning algorithms inspired by the human brain's network of neurons \citep{goodfellow2016deep}. These algorithms use a hierarchical structure of neural layers to extract higher-level features from the raw input progressively. In the past few years, the growing computational power of modern GPU-based computers and the availability of large training datasets in the field of machine learning have made it possible to successfully train neural networks with many layers and degrees of freedom. Consequently, deep learning has revolutionized many machine learning tasks in recent years, ranging from image and video processing to speech recognition and natural language understanding.

Many neuroscientific research results served as focal points in the development of deep learning algorithms. When \citet{hubel1962receptive} studied the visual cortex in the brain, they made three important discoveries. First, they observed a one-to-one correspondence between spatial locations in the retina and neurons in the brain that fired as a response to line-like visual stimuli. Second, the activity of the neurons changed depending on the orientation of the line, uncovering a neat organization based on local orientations. Last, the neurons sometimes fired only when the line was moving in a particular direction. Later, \citet{bosking_orientation_1997} showed that neurons that are aligned fire together, indicating the presence of a type of long-range interactions. All these results motivated the development of a mathematical framework for modeling visual perception based on sub-Riemannian geometry on the space of positions and orientations, which is typically modeled with the Lie group SE(2) \citep{petitot_neurogeometry_2003,citti_cortical_2006,duits_association_2014}. Apart from the neurophysiological inspiration, group equivariance has also been proven to be an excellent inductive bias \citep{cohen2016gcnn} not only in computer vision (as the translation equivariance property of CNNs as shown) but also in physics \citep{finzi2020generalizing} and molecular data analysis \citep{fuchs2021iterative,
jumper2020high}. In this work, we propose to build group equivariant graph neural networks via the same principle that underlie the sub-Riemannian, neurogeometrical modeling of the visual cortex.

Our work connects the observations by \cite{hubel1962receptive} and \cite{bosking_orientation_1997} on two levels. First, the organization of visual data based on their location and orientation \citep{hubel1962receptive} is modeled by Lie group convolutions \citep{bekkers2019b}, in which feature maps encode response for every position and every orientation. Second, long-range interactions between aligned neurons \citep{bosking_orientation_1997} are modeled by building graphs with affinity matrices based on (approximate) sub-Riemannian distances on the Lie groups, inspired by sub-Riemannian image analysis methods such as \citep{franken_crossing-preserving_2009,bekkers_pde_2015,favali_analysis_2016,mashtakov2017tracking,boscain2018highly,duits_optimal_2018,baspinar2021cortical}. 

\citet{defferrard2020deepsphere} showed how to construct powerful graph NNs that are faithful to the manifolds on which they are defined. Nevertheless, the layers themselves are based on rotationally invariant (Laplacian) convolutions. In order to exploit directional cues in the data, group convolutions are desirable \citep{cohen2018general, kondor2018generalization, cohen2016gcnn,bekkers2019b}. However, since Laplacian operators are intrinsically isotropic, there is no point applying them to the lifted feature maps on the group unless we  construct anisotropic metrics on the groups. Therefore, we adopt the Lie group viewpoint by \citet{sanguinetti2015fastmarching} to define anisotropic Riemannian metrics based on left-invariant vector fields on the group. Once an anisotropic Riemannian graph is constructed, any spectral method can directly be applied to this graph. The resulting graph neural networks will then, by construction, be equivariant and capable of utilizing directional cues in data.

Before going further into the details, we summarize our main contributions:
\begin{itemize}
\item We introduce ChebLieNet, an equivariant graph Laplacian-based neural network based on Lie groups equipped with an anisotropic Riemannian metric. 
\item The Riemannian geometry is automatically derived from a standard base space (e.g. $\mathbb{R}^2$ or the sphere), which makes our approach flexible and effective in building group equivariant graph neural networks for a variety of data structures (e.g. 2D and spherical data).
\item We demonstrate the equivariance property of ChebLieNet, both in theory and in practice. This property guarantees that the neural network's predictions are robust against given transformations, which is not necessarily the case with methods based on data augmentation.
\item We show that the use of directional information via anisotropic Riemannian spaces could benefit many tasks. 
\item We show the flexibility of the method by considering two different problems; we validate on classification problems with 2D image data and a segmentation problem on spherical data via the construction of a sub-Riemannian geometry on $SE(2)$ and $SO(3)$ respectively.
\end{itemize}

\section{Related works} \label{sec:related_works}

\subsection{Group equivariant convolutional neural networks} \label{group_equivaiant_convolutional_neural_networks}

Deep convolutional neural networks \citep{lecun1995convolutional} have proven to be compelling models for pattern recognition tasks on images, video, and audio data. Although a robust theory of neural network design is currently lacking, a large amount of empirical evidence supports the notion that both convolutional weight sharing, depth, and width are essential for good predictive performance. Such properties are enabled through the equivariance property of convolutions (convolving a shifted image is the same as translating its result).

\citet{lenc2015understanding} showed that the AlexNet CNN \citet{krizhevsky2012imagenet} trained on ImageNet learns representations equivariant to flips, scalings, and rotations spontaneously. This supports the idea that equivariance is an excellent inductive bias for deep convolutional networks. In the last few years, a joint effort has been made to build group equivariant networks. By the introduction of group convolutions in deep learning, \citet{cohen2016gcnn} generalize the translation equivariance property to larger groups of symmetries, including rotations and reflections. \citet{kondor2018generalization} gave a rigorous, theoretical treatment of convolution and equivariance in neural networks concerning any compact group's action.  One of the main contributions of that work was to show that, given some natural constraints, the convolutional structure is not just a sufficient but also a necessary condition for equivariance to a compact group's action. In a similar spirit, in \citep{bekkers2019b} it is shown that any bounded linear operator is equivariant to Lie groups if and only if it is a group convolution. In our work, we propose to build group equivariant neural networks via left-invariant Laplace operators on Lie groups, which indeed can be seen as group convolutions with kernels that are the fundamental solutions of the Laplace operator. The result is a Lie group equivariant Chebyshev-type neural network \citep{defferrard2016chebnet} that we will refer to as ChebLieNet.

\subsection{Graph neural networks} \label{sec:graph_neural_networks}

Using the term geometric deep learning, \citet{bronstein2017geometric,bronstein2021geometric} give an overview of deep learning methods in the non-Euclidean domain, including graphs and manifolds. They present different examples of geometric deep learning problems and available solutions, fundamental difficulties, applications, and future research directions in this nascent field.

One of the main challenges when working with graph data it to deal with the inter-dependencies between points. Indeed, the derivations of most standard machine learning models firmly base on an independence assumption. For this reason, transferring existing methods on a graph appears doomed to failure, and it seems necessary to build models acting directly on graphs. Due to its success on Euclidean data, the development of a convolution-like operator on graphs has been largely studied. Because the notion of space is not naturally defined on a graph, we lack a straightforward generalization of the convolutional operator from grid data to graphs \citep{scarselli2008graph, bruna2013spectral, henaff2015deep, defferrard2016chebnet, kipf2016gcn, masci2015geodesic, boscaini2016learning, monti2017geometric}.

Spectral approaches have a solid mathematical foundation in graph signal processing. Rather than using the traditional spatial definition of the convolution, it proposes to see this operation from a spectral perspective. Based on the convolution theorem, it defines the convolution operator from the graph spectral domain via the eigendecomposition of the graph Laplacian (see App.~\ref{app:graph_theory}).

\begin{definition}[Spectral graph convolution]
Let $\mathcal{G} = (\mathcal{V}, \mathcal{E}, \boldsymbol{W})$ be a graph with Laplacian $\boldsymbol{\hat{\Delta}}$ and let $f$ and $g$ be two functions defined on $\mathcal{V}$. We define the $\mathcal{G}$-convolution $*_{\mathcal{G}}$ of $f$ and $g$ as:
\begin{equation}
f *_{\mathcal{G}} g = \boldsymbol{\Phi} (\boldsymbol{\hat{g}} \odot \boldsymbol{\hat{f}}) = \boldsymbol{\Phi} (\boldsymbol{\Phi}^{\top} \boldsymbol{g} \odot \boldsymbol{\Phi}^{\top} \boldsymbol{f}),
\end{equation}
with eigenvectors $\boldsymbol{\Phi}$ obtained through the unique eigendecomposition $\boldsymbol{\hat{\Delta}} = \boldsymbol{\Phi} \boldsymbol{\Lambda} \boldsymbol{\Phi}^T$.
\end{definition}

While this definition alleviates the difficulty of deriving a convolution operator in the spatial domain, other difficulties arise. First of all, because the Laplacian of a graph is an intrinsic operator, it is domain-dependent, and the spectral-convolution is too. It implies that a model built on this framework cannot be easily transferred from a graph to another as expressed in a different "language". Nevertheless, this is not a problem for us since we are focusing on fixed manifold graphs. Next, there is no guarantee that filters represented in the spectral domain are spatially localized. \citet{henaff2015deep} successfully bypassed this problem by defining smooth spectral filter coefficients, arguing that if spectral filters are smooth, they are spatially localized. Last but not least, the Laplacian's eigendecomposition makes the method expensive in terms of memory and time. Indeed, the forward and inverse graph Fourier transforms (via $\boldsymbol{\Phi}^T$ and $\boldsymbol{\Phi}$) incur expensive multiplications as no FFT-like algorithm exists on general graphs. \citet{defferrard2016chebnet} alleviated the cost of explicitly computing the graph Laplacian using spatially-localized filters with Chebyshev polynomials.

\begin{definition}[Chebyshev convolutional layer] \label{def:cheb_conv}
Let $\mathcal{G} = (\mathcal{V}, \mathcal{E}, \boldsymbol{W})$ be a graph with rescaled Laplacian\footnote{Because Chebyshev polynomials are defined in the range $[-1, 1]$, it is necessary to rescale the graph Laplacian with $\boldsymbol{\tilde{\Delta}} = 2\lambda_{\max}^{-1}\boldsymbol{\hat{\Delta}} - \boldsymbol{I}$ where $\lambda_{\max}$ is the largest eigenvalue of $\boldsymbol{\hat{\Delta}}$.} $\boldsymbol{\tilde{\Delta}}$, $\boldsymbol{x} \in \mathbb{R}^{|\mathcal{V}| \times d_i}$  be an input features' vector and $\Theta_j \in \mathbb{R}^{d_i \times d_o}$ learnable filters. The output features' vector $\boldsymbol{y} \in \mathbb{R}^{|\mathcal{V}| \times d_o}$ is computed as:
\begin{equation}
\boldsymbol{y} = \sum_{j=0}^{R-1} \boldsymbol{z}_j \boldsymbol{\Theta}_j \qquad \text{with} \quad \boldsymbol{z}_0 = \boldsymbol{x}, \quad \boldsymbol{z}_1 = \boldsymbol{\tilde{\Delta} x} \quad  \text{and} \quad \boldsymbol{z}_j = 2 \boldsymbol{\tilde{\Delta} z}_{j-1} - \boldsymbol{z}_{j-2}. \quad \forall j \geq 2.
\end{equation}
\end{definition}

\citet{kipf2016gcn} simplified this formulation a bit by considering the construction of single-parametric filters that are linear with relation to $\boldsymbol{\tilde{\Delta}}$. They further approximate $\lambda_{\max} \simeq 2$ as they expect that neural network parameters will adapt to this change in scale during training.

\section{Method} \label{sec:method}

Our method can be seen as an extension of the original ChebNet \citep{defferrard2016chebnet, perraudin2019deepsphere}. Instead of directly working on a homogeneous base space, we first extend it to a higher dimensional space (Lie group). The goal of this extension is to convert the previously invariant spectral convolutional layers into equivariant layers.\footnote{Because spectral graph NNs are able to capture the geometry of the space, which in this work we equip with anisotropic metrics, any spectral method could be made equivariant using our method.}

\subsection{Anisotropic manifold graph} \label{sec:anisotropic_manifold_graph}

In order to define the anisotropic manifold graphs we have to consider two types of manifolds. The base manifold $\mathcal{M}$ and a Lie group $G$ that acts transitively on $\mathcal{M}$. The latter implies that $\mathcal{M}$ is a homogeneous space of $G$, which means that any two points $m_1, m_2 \in \mathcal{M}$ can be mapped to each other via the action of a group element $g \in G$ via $m_2 = g \cdot m_1$. E.g., the plane $\mathcal{M}=\mathbb{R}^2$ is a homogeneous space of the special Euclidean motion group $G=SE(2)$ as any two points can be mapped to each other through a rotation and a translation. Such groups $G$, which have $\mathcal{M}$ as a homogeneous space, can always be split in two parts via the semi-direct product $G = \mathcal{M} \rtimes H$, with $H$ a sub-group of $G$ that leaves some reference point $m_0 \in \mathcal{M}$ invariant, i.e., $\forall_{h \in H}: \,m_0 = h \cdot m_0$. E.g., rotations leave the zero vector in $\mathcal{M}=\mathbb{R}^2$ invariant, and thus $H=SO(2)$ in the $SE(2)$ case. Conversely, any homogenous space can be modeled with a group quotient $\mathcal{M} = G / H$.

We define an anisotropic manifold graph to be a discretization of a Lie group $G$ of which $\mathcal{M}$ is a homogeneous space. It consists of a finite set of vertices corresponding to a random sampling of group elements, and a finite set of similarity-based edges that are constructed via a left-invariant Riemannian metric on $G$. In our work we consider two anisotropic manifold graphs: one associated with the base manifold $\mathcal{M} = \mathbb{R}^2$ which we extend with an additional orientation/rotation dimension $H = SO(2)$ to come to the Lie group $G = SE(2) = \mathbb{R}^2 \rtimes SO(2)$, and the other associated with the sphere $\mathcal{M} = S^2$ which we similarly "lift" to the Lie group $G = SO(3)$ by adding an additional rotation dimension. 
Considering the similarity between the two cases (the sphere locally looks like $\mathbb{R}^2$) we will refer to $\mathcal{M}$ as the "spatial" part, and $H$ as the "orientation" part of the group. 

\paragraph{Uniform sampling of the vertices.} The first step to construct an anisotropic manifold graph is to sample elements on the group uniformly or as uniformly as possible if the manifold does not permit a uniform grid. We split the grid construction in two parts, a grid on $\mathcal{M}$ which is sampled with $|\mathcal{V}_s|$ points and a grid on $H$ that is sampled with $|\mathcal{V}_o|$ points, leading to a total of $|\mathcal{V}| = |\mathcal{V}_s| |\mathcal{V}_o|$ vertices.

\paragraph{Left-invariant anisotropic Riemannian distance.} Once vertices have been uniformly sampled on the group manifold, a similarity measure between vertices is computed. This measure is based on a Riemannian distance between points in $G$. The only thing one needs in our algorithm is the implementation of the logarithmic map on the Lie group (see e.g. \citep{bekkers2019b}), and a diagonal Riemannian metric tensor (see e.g. \citep{sanguinetti2015fastmarching} and \citep{mashtakov2017tracking} for the $SE(2)$ and $SO(3)$ case respectively). In the following we provide the essential idea and intuition behind the construction of the similarity measure and provide a more extensive treatment in App.~\ref{app:lie_groups}.

In Riemannian geometry on Lie groups it is common to express tangent vectors of curves in a basis of left-invariant vector fields as it allows to measure their lengths with a single Riemannian metric tensor that is shared over the entire group. This works as follows. Consider curve $\gamma:[0,1]\rightarrow G$ with its tangent vectors $\dot{\gamma}(t) = \sum_{i=1}^d u^i(t) \mathcal{A}_i|_{\gamma(t)}$ expressed in a basis/moving frame of reference $\{\mathcal{A}_i|_{\gamma(t)}\}_{i=1}^d$, in which $\mathcal{A}_i$ are left-invariant vector fields. The length of these tangent vectors is then measured by a Riemannian metric tensor that we denote with $\lVert \dot{\gamma}(t) \rVert_{\mathbf{R}}^2 := \mathbf{u}(t)^T \mathbf{R} \mathbf{u}(t)$, with $\mathbf{R}$ a symmetric positive definite matrix defined relative to the basis $\{\mathcal{A}_i|_{\gamma(t)}\}_{i=1}^d$, and with $\mathbf{u}(t) = (u_0(t), u_1(t), \dots )^T$. The $\mathcal{A}_i$ are left-invariant vector fields and the notation $\mathcal{A}_i|_{g}$ means the vector in the vector field $\mathcal{A}_i$ at location $g$. The vector fields are constructed by choosing a vector $A_i$ in the tangent space at origin (the Lie algebra) which then defines a complete vector field on $G$ via the push-forward of left-multiplication. In less technical terms this means that if we pick a direction vector at the origin, and we move it to another point in, e.g. $G=SE(2)$, via a roto-translation, this vector will move and rotate along. By defining everything in terms of these left-invariant vector fields, every tangent space $T_g(G)$ at each $g \in G$ can be identified with the tangent space at the origin, and a single Riemannian metric tensor $\mathbf{R}$ can be shared over the entire space. Moreover, the induced Riemannian distance $d(g,h)$ between any two points $g,h \in G$ is then by construction left-invariant, i.e., $\forall_{g,h,i \in G}:\, d(g \cdot h, g \cdot i) = d(h,i)$.

Expressing tangent vectors in such left-invariant vector fields allows us to reason in terms of the generators of the group and even design learnable equivariant differential operators for the construction of deep NNs \citep{smets2020pde}. Consider the $G=SE(2)$ case. As a basis we pick the 3 generators of the group: a forward motion represented by a vector $A_1$ pointing in the forward direction within the plane, a side-ways motion represented by a perpendicular planar vector $A_2$, and a rotation/change of orientation represented by a vector $A_3$ that points vertically in along the $H$-dimension. We then work with diagonal Riemannian metric tensors $\mathbf{R} = \diag (1, \epsilon^{-2}, \xi^{2})$, which penalize each type of motion (represented by the vector components) differently. When $\epsilon \rightarrow 0$ one arrives at the \textit{sub-Riemannian geometry} which forms the basis for the mathematical modeling of visual perception. It quantifies a notion of alignment through the sub-Riemannian distance; the length of a distance-minimizing geodesic that connects two local orientations that lie in the extend of each other will be much smaller that that of a geodesic connecting two local orientations parallel to each other. An analogy can be found with the example of a car in a parking lot where it can move forward/backward ($A_1$) and change orientation ($A_3$) \citep{reeds1990optimal}. It will be easier to move it to the more aligned spot directly ahead then it will to the spot next to the car, as sideways motion ($A_2$) is impossible.  

\begin{figure*}[h!]
    \centering
    \begin{subfigure}[b]{0.3\textwidth}
        \centering
        \includegraphics[width=\textwidth]{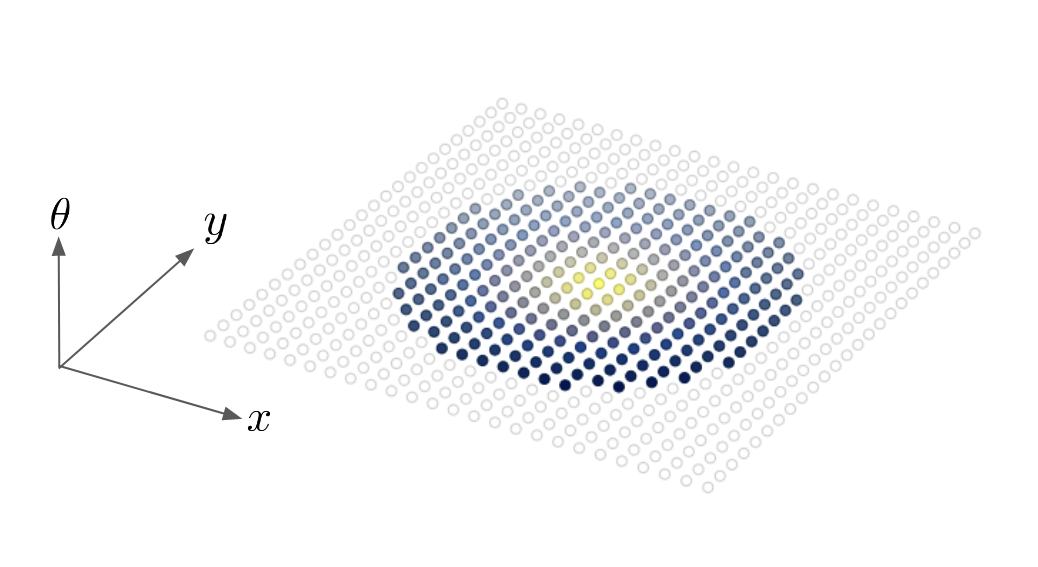}
        \caption{Base space $\mathcal{M}$ with an isotropic Riemannian metric}
    \end{subfigure}
    \hfill
    \begin{subfigure}[b]{0.3\textwidth}
        \centering
        \includegraphics[width=\textwidth]{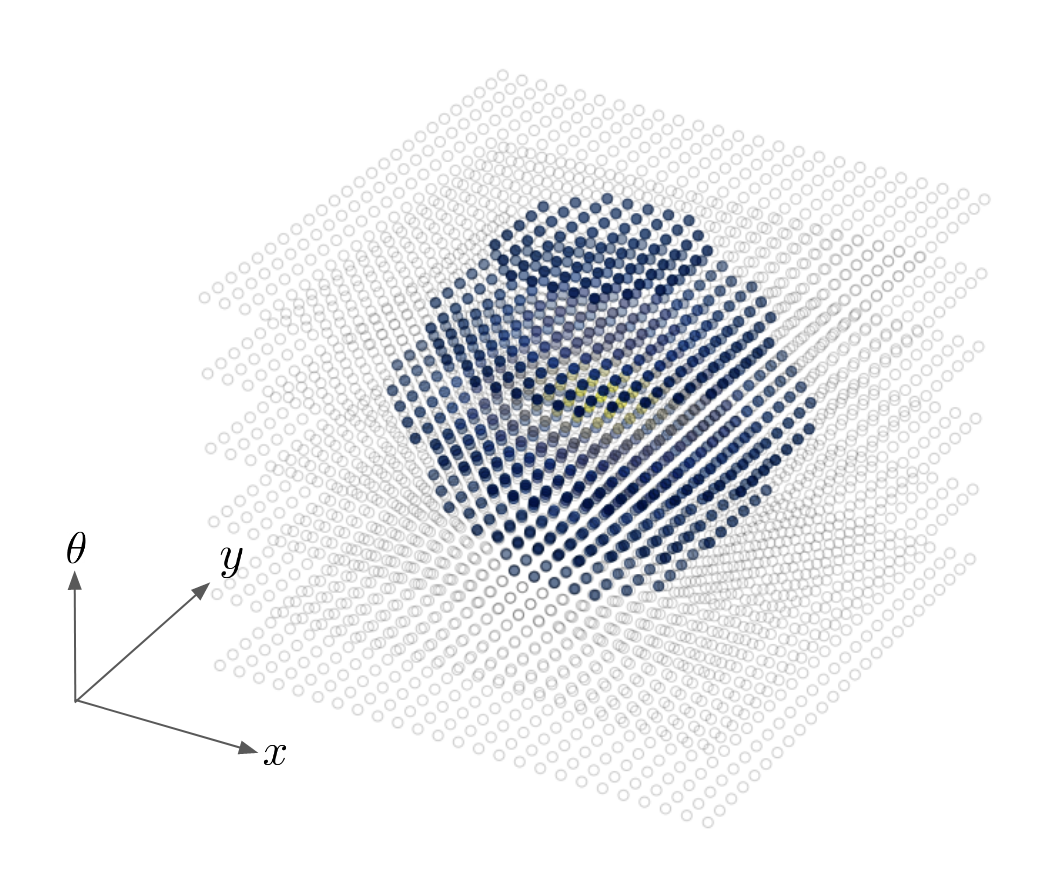}
        \caption{Lie group extension $G$ with an isotropic Riemannian metric}
    \end{subfigure}
    \hfill
    \begin{subfigure}[b]{0.3\textwidth}
        \centering
        \includegraphics[width=\textwidth]{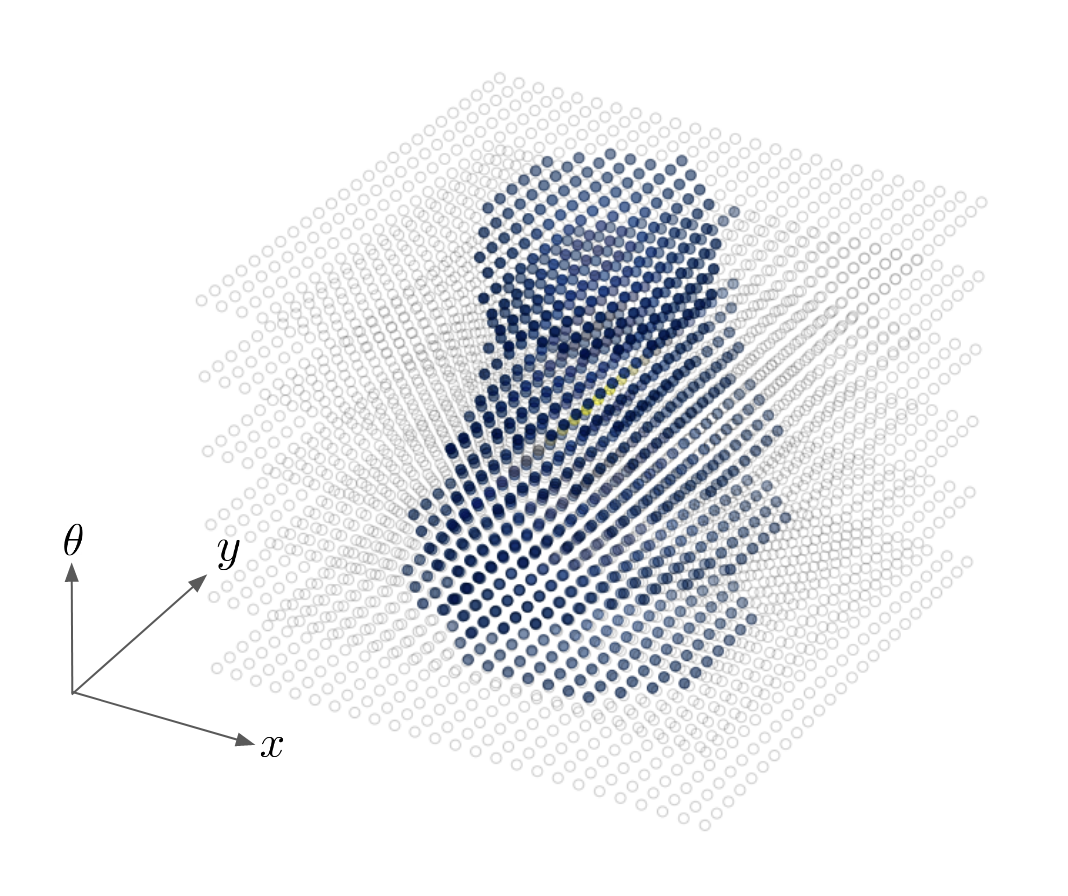}
        \caption{Lie group extension $G$ with an anisotropic Riemannian metric}
    \end{subfigure}
    \caption{Isotropic diffusion applied to an impulse signal on Riemannian manifolds on $\mathcal{M}=\mathbb{R}^2$ and $G=SE(2)$. }
    \label{fig:graph_diffusion}
\end{figure*}

Parameters $\epsilon$ and $\xi$ will respectively be referred to as spatial and orientation anisotropy parameters. With $\epsilon=1$ the metric is isotropic and there will be no distinction between different orientations. When $\epsilon < 1$, $\xi$ determines the flexibilty/curvature of the geodesics as it balances spatial motion against angular motion. In a sense it defines how easily one connects local orientations that are not optimally aligned. In Figure \ref{fig:graph_diffusion} this behavior is visualized by running a diffusion process on the anisotropic manifold graph. In the anisotropic case ($\epsilon<1$) diffusion is faster along the forward direction within a $\theta$-plane. From a graph NN perspective this suggests that information is propagated more quickly between vertices that are aligned, nevertheless, Chow's theorem (see e.g. \citep{mont}) guarantees that any point pair in the (sub-)Riemannian manfiold can interact with one another.

The exact computation of the (sub-)Riemannian distances is challenging and can generally not be done in closed form, but can be done numerically via method such as \citep{bekkers_pde_2015,sanguinetti2015fastmarching,mashtakov2017tracking}. In order to keep our graph construction algorithm efficient though, we will approximate the Riemannian distances via an efficient analytic formula based on those in \citep{bekkers2018nilpotent} that only involves the Lie group's logarithmic map $\log:G \rightarrow T_e(G)$ and the Riemannian metric tensor $\mathbf{R}$. We then approximate the distance between points $g,h \in G$ by 
\begin{equation}
d(g, h) = d(e, g^{-1} \cdot h) \simeq ||\log (g^{-1} \cdot h)||_{\mathbf{R}}.
\end{equation}


\paragraph{Similarity measure.} Encoding a similarity measure in the edges of a graph requires defining a weighting scheme. It is common to use a Gaussian kernel and set the weights via
\begin{equation}
w(v_i, v_j) =
\left\{
\begin{array}{ll}
\exp \left(- \frac{d^2(v_i, v_j)}{4t}\right) & \text{if } e(v_i, v_j) \in \mathcal{E} \\
0 & \text{otherwise}
\end{array}
\right. .
\end{equation} 
The choice for kernel bandwidth $t$ is essentially arbitrary, but good heuristics exist. \citet{perraudin2019deepsphere} set it to half the average squared distance between connected vertices. \citet{defferrard2020deepsphere}, however, showed that this heuristic has the tendency to overestimate it and preferred to choose it as the minimizer of the mean equivariance error. Following this overestimation observation, we fix the kernel bandwidth as $20\%$ of the average squared Riemannian distance between connected vertices. As such, the weights diversely cover values in the whole range $[0, 1]$. The most similar vertices are connected with close-to-one weighted edges whereas the lowest connections are close to zero.

\paragraph{Quality of the approximation.} In theory, we would like our approximation to be as precise as possible. In practice, a high-resolution approximation leads to computational issues in time and memory. Hence, tuning of the graph parameters becomes a trade-off between theoretical consistency and practical feasibility. First of all, the graph resolution (or the number of vertices we sample) is directly related to the quality of the approximation. While the spatial resolution $|\mathcal{V}|_s$ is usually determined by the data (up to up- and down-samplings), the orientation resolution $|\mathcal{V}|_o$ is a design choice. An important remark is to notice that a large orientation resolution does not necessarily help if two different orientations are not distinguishable because of a poor spatial resolution \citep{weiler_learning_2018,bekkers2019b}. 
Secondly, the connectivity of the graph is also a crucial parameter. A fully connected graph is theoretically the best approximation. Nevertheless, for computational reasons, we use $K$-NN graphs\footnote{Note that in our implementation, a $K$-NN graphs does not mean that each vertex has $K$ neighbors but at most $K$ neighbors. Indeed, if the graph domain has boundaries, using exactly $K$ neighbors for each vertex could lead to asymmetries that may introduce biases and harm the permutation invariances in the graph.} to sparsify the graph Laplacians.

\paragraph{Theoretical group equivariance of the graph Laplacian.} Due to the success of machine learning algorithms based on graph Laplacian, the theoretical convergence of the graph Laplacian to its continuous analogue has been largely studied \citep{hein2005graphs, singer2006graph}. \cite{belkin2006convergence} noticed that in many graph-based algorithms, a central role is played by the graph Laplacian's eigenvectors. Thus, they focused on proving convergence in eigenmaps as it is sufficient in this case. They proved that if the graph's vertices are sampled uniformly from an unknown submanifold $\mathcal{M} \in \mathbb{R}^d$, then the eigenvectors of a suitably constructed graph Laplacian converges to the eigenfunctions of the Laplace-Beltrami operator on $\mathcal{M}$. Consequently, as the latter operator is left-invariant, as we show in theorem \ref{thm:equiv_laplace_beltrami}, the graph Laplacian is asymptotically \footnote{The asymptotic case corresponds to $|\mathcal{V}| \to \infty$ and a Gaussian weight kernel with kernel bandwidth $t \to 0$.} group equivariant.

\paragraph{Empirical group equivariance of the graph Laplacian.} We empirically confirm the group equivariance property of the graph Laplacian applied to our anisotropic manifold graphs. By checking $\boldsymbol{P^\top \tilde{\Delta} P} = \boldsymbol{\tilde{\Delta}}$ where $\boldsymbol{P}$ is a permutation matrix, we can verify that the graph Laplacian is invariant under a given permutation of vertices corresponding to a group transformation (e.g. a rotation of the graph). Moreover, we can also compare the eigenmaps of a graph Laplacian and its continuous counterpart if it is well-known. For a further discussion about this, see App.~\ref{app:laplacian}.

\subsection{ChebLieNet} \label{cheblienet}

\begin{SCfigure}
    \centering
    \includegraphics[width=0.7\textwidth]{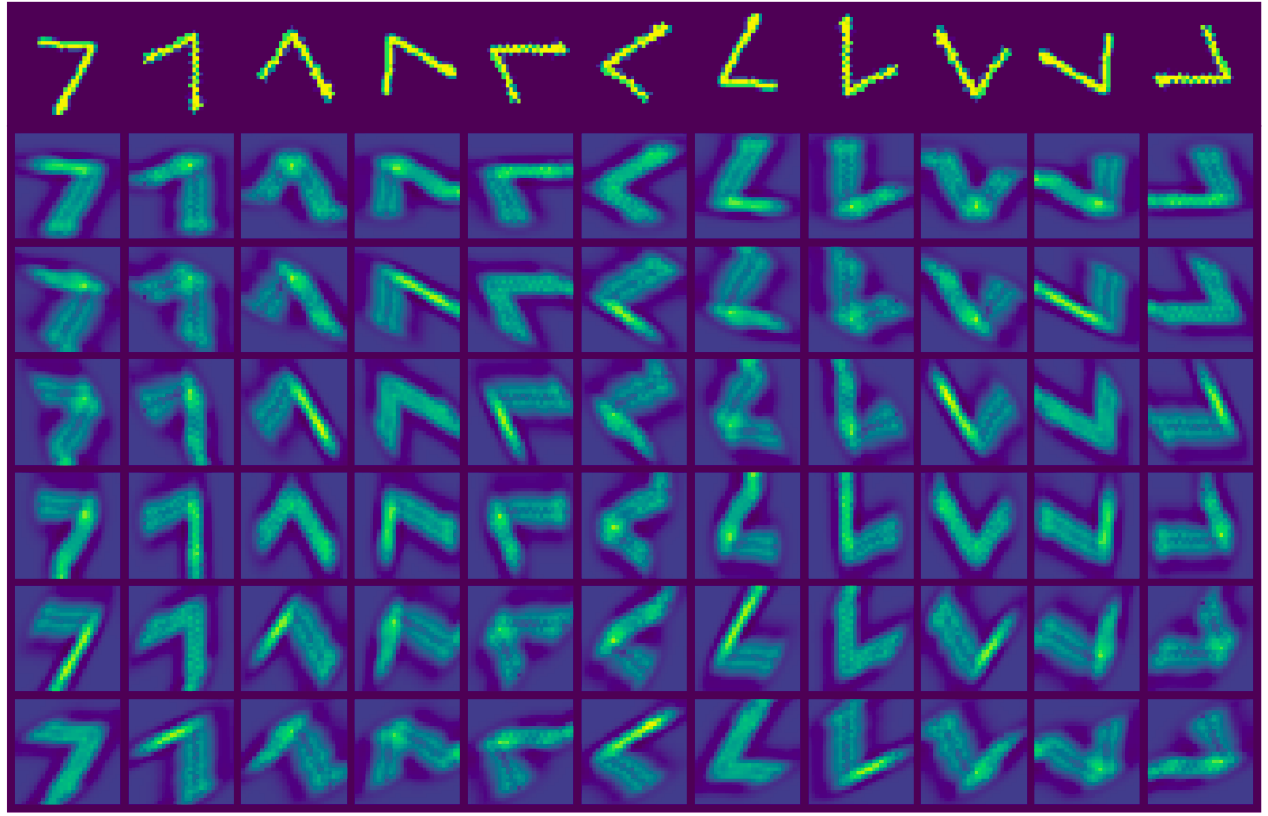}
    \caption{Rotation equivariance of a randomly initialized $SE(2)$ Chebyshev convolutional layer. From left to right shows different rotations of an input (top row) and the activations for different slices of $\theta \in [0,\pi]$ in the graph (bottom 6 rows). A rotation of an input image followed by Chebyshev convolution is equivalent to first convolution followed by a planar rotation in each $\theta$ slice and a roll in the $\theta$-axis.}
    \label{fig:7_filters}
\end{SCfigure}

\paragraph{Chebyshev convolutional layer.} As introduced in \citet{defferrard2016chebnet}, a Chebyshev convolutional layer is a spectral layer based on a continuous kernel parametrization with graph Laplacians. This parameterization makes such layers highly suitable for our method, as they intrinsically capture the Riemannian geometry of the graphs on $G$. 
Moreover, the Chebyshev convolutions on the anisotropic manifold graphs are equivariant by construction because the graph Laplacians are equivariant operators (see Figure~\ref{fig:7_filters}).

\begin{figure*}[h!]
    \centering
    \begin{subfigure}[b]{0.48\textwidth}
        \centering
        \includegraphics[width=\textwidth]{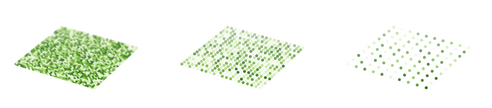}
        \caption{R2RandPool}
    \end{subfigure}
    \hfill
    \begin{subfigure}[b]{0.48\textwidth}
        \centering
        \includegraphics[width=\textwidth]{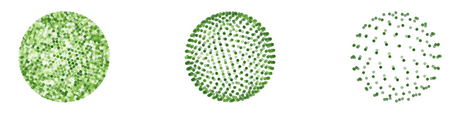}
        \caption{S2MaxPool}
    \end{subfigure}
    \hfill
    \begin{subfigure}[b]{0.48\textwidth}
        \centering
        \includegraphics[width=\textwidth]{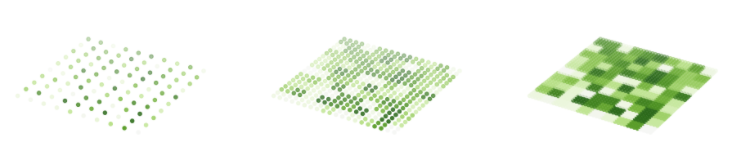}
        \caption{R2RandUnpool}
    \end{subfigure}
    \hfill
    \begin{subfigure}[b]{0.48\textwidth}
        \centering
        \includegraphics[width=\textwidth]{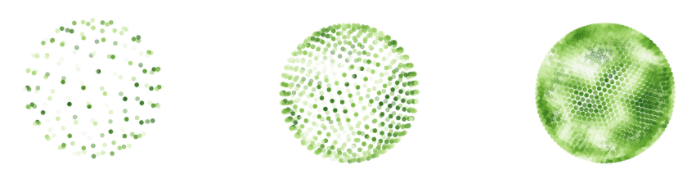}
        \caption{S2AvgUnpool}
    \end{subfigure}
    \caption{Spatial pooling and unpooling layers on the 2D grid and the sphere.}
    \label{fig:pooling}
\end{figure*}

\paragraph{Spatial pooling and unpooling layers.} Graph pooling is a central component in a myriad of graph neural network architectures. Producing coarsened graphs from a finer graph have two main advantages: first, it reduces the computational cost, and second, it could improve performance by reducing the overfitting effect and adding a multiscale perspective. As an inheritance from traditional CNNs, most approaches formulate graph pooling as a cluster assignment problem, extending local patches' idea in regular grids to graphs \citep{dhillon2007weighted, ying2018hierarchical, khasahmadi2020memory, mesquita2020rethinking}. We propose similar operations on the base space (spatial domain) and involving two steps (see Figure~\ref{fig:pooling}). First, each sample is assigned to a cluster that will correspond to the output sample; this is the down- (resp. up-) sampling phase. With a well designed method, this change of data-resolution can be made equivariant to any group transformation.\footnote{Altough down- and up-samplings are naturally defined on the Euclidean grid, this task is more complicated on the sphere. However, using an icosahedron decomposition of the sphere, we make it more natural as down- and up-sampling consists of decreasing or increasing the subdivision level.} Then, each cluster is reduced (resp. expanded) according to a given scheme (e.g. maximum, average or random); this is the reduction (resp. expansion) phase. When the reduction and expansion steps are permutation-invariant operations, such layers are automatically invariant under any transformation in the group. 

\paragraph{Global pooling (projection) layer and point-wise operations.} When the neural network does not need to be equivariant but invariant (e.g. classification task), it is common to rely on a global pooling layer (or simply projection layer). This layer reduces the d-dimensional signal on the graph's vertices to a d-dimensional vector of features derived from information on the whole graph. As a permutation-invariant operation, such a layer does not break the equivariance property of the neural network. Finally, point-wise operations do not affect the equivariance of a neural network. 

\section{Experiments} \label{sec:experiments}

In this section, we show the benefits of working on the anisotropic manifold graphs compared to the base manifold graphs. We believe that further improvements could be achieved through tuning and hyper-parameter optimization of the models \citep{yu2020hyper}, using high-capacity networks, or via a more advanced training process, but this is not the goal of our work. We here intent to illustrate the adaptability of our approach to different tasks such as classification and segmentation in 2D images or spherical data. In the first couple of experiments, we motive the use of anisotropic spaces. By varying the anisotropies, we show the existence of sweet spots, both for the spatial anisotropy parameter $\epsilon$ and the orientation anisotropy parameter $\xi$. In the second couple of experiments, we show that even if we add a new orientation dimension, our method remains scalable using a proper implementation.

Our implementation is fully PyTorch \citep{pytorch} and available at \url{https://github.com/haguettaz/ChebLieNet}.
We perform all the experiments on a single GeForce GTX 1080 Ti gpu and track them with the Weights \& Biases library \citep{wandb}. The details of the experiments are given in the App.~\ref{app:experiment_details}.

\subsection{Why using tunable anisotropic kernels?}

As introduced in Section~\ref{sec:anisotropic_manifold_graph}, the anisotropies are tunable via the parameters $\epsilon$ and $\xi$ of the Riemannian metric, respectively responsible for the spatial and orientation anisotropies. As the $\xi$ parameter should depend on the spatial and orientation resolutions, we use the following parameterisation: $\xi^2 = \alpha \frac{|\mathcal{V}_o|}{|\mathcal{V}_s|}$. Setting $\alpha = 1$ yields a 40/60 ratio of neighbors within versus outside the orientation plane. We ran different experiments with a Wide Residual architecture \citep{zagoruyko2016wide} on CIFAR10 \citep{krizhevsky2009learning}, varying the spatial and orientation anisotropic parameters.

\begin{figure*}[h!]
    \centering
    \begin{subfigure}[b]{0.48\textwidth}
        \centering
        \begin{tikzpicture}
        \begin{axis}[
            xlabel={$\alpha$},
            xmode=log,
            log ticks with fixed point,
            ylabel={Test-accuracy [\%]},
            xmin=1, xmax=32,
            ymin=77, ymax=87,
            xtick={1,2,4,8,16,32},
            ytick={80, 85},
            legend pos=north east,
            ymajorgrids=true,
            grid style=dashed,
            height=4cm,
            width=6cm
        ]
        \addplot[
            color=color1,
            mark=square,
            ]
            coordinates {
            (1,83.43)(2,84.83)(4,83.41)(8,78.85)(16,80.06)(32,80.46)
            };
        \end{axis}
        \end{tikzpicture}
        \caption{Test-accuracy against orientation anisotropies}
        \label{fig:sweet_spot_orientation}
    \end{subfigure}
    \hfill
    \begin{subfigure}[b]{0.48\textwidth}
        \centering
        \begin{tikzpicture}
        \begin{axis}[
            xlabel={$\epsilon^2$},
            ylabel={Test-accuracy [\%]},
            xmin=0.1, xmax=1.0,
            ymin=75, ymax=87,
            xtick={0,0.2,0.4,0.6,0.8,1.0},
            ytick={80, 85},
            legend pos=north east,
            ymajorgrids=true,
            grid style=dashed,
            height=4cm,
            width=6cm
        ]
        \addplot[
            color=color1,
            mark=square,
            ]
            coordinates {
            (0.1,84.26)(0.2,85.49)(0.5,84.69)(1.0,77.34)
            };
        \end{axis}
        \draw[stealth-,  line width=0.6pt] (4.4,0.5) -- (3.5,1.5) node[above, align=center]{\small isotropic \\ \small \& invariant};
        \draw[stealth-, line width=0.6pt] (0.1,0.2) -- (4.2,0.2);
        \draw (2.1,0.4) node[align=center]{\small equivariant};
        \end{tikzpicture}
        \caption{Test-accuracy against spatial anisotropies}
        \label{fig:sweet_spot_spatial}
    \end{subfigure}
    \caption{Empirical proof of existence of sweet spots for data-dependent anisotropic parameters.}
\end{figure*}
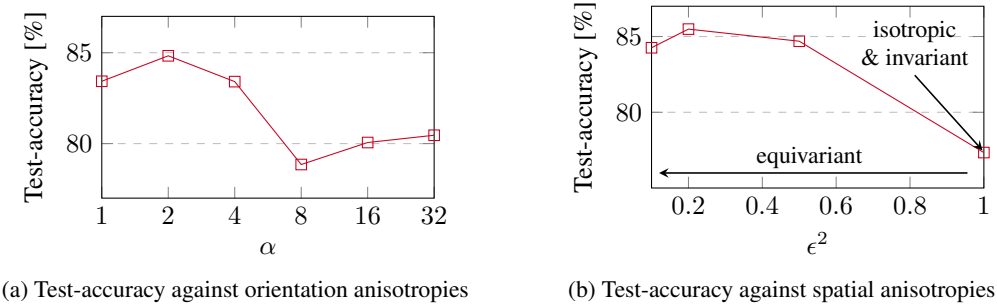

\paragraph{Orientation anisotropy.} The orientation anisotropy $\xi$ controls how strongly orientation layers are connected. At the limit $\xi \to \infty$, orientation layers are decoupled. It is like test-time augmentation with rotations: running a CNN working with one anisotropic Laplacian (e.g., only vertically aligned filters) and testing the network for different input rotations before averaging the output. The other extreme $\xi \to 0$ keeps all layers equally close to each other, and features are essentially identified with just a spatial coordinate. This would then correspond to a WideResNet with isotropic Chebyshev convolutions. For reasonable values of $\xi$, interactions between orientation layers take place. Figure \ref{fig:sweet_spot_orientation} is evidence of the existence of a sweet spot for this parameter in the range of reasonable values. At the moment, we expect with no certainty that this parameter could be set \emph{a priori} of the data, only considering the data resolution. As a rule of thumb, we set $\xi$ such that each vertex has approximately 40\% of its neighbors in the same orientation layer and 60\% on others.

\paragraph{Spatial anisotropy.} The spatial anisotropy $\epsilon$ regulates the anisotropy of the space on the spatial domain. For $\epsilon = 1$, the Riemannian metric is spatially isotropic; all directions are treated equally and the resulting model would effectively be a WideResNet with isotropic Chebyshev convolutions. At the limit $\epsilon \to 0$, the main direction has a minimal cost, and the resulting space is highly spatially anisotropic. In figure \ref{fig:sweet_spot_orientation} we observe that using anisotropic spaces instead of isotropic ones is relevant, as we almost get an 8\% test-accuracy improvement. Unlike the orientation anisotropic parameter, in our opinion, this parameter is task/data-dependent; different datasets could benefit in different degrees from the utilization of directional information through different spatial anisotropy settings.

\subsection{How scalable is the method?}

Scalability is often an important limitation of graph- and group-based neural networks. By adding an orientation dimension, we do not run from this rule as we necessarily increase the number of vertices of the anisotropic manifold graphs. To permit experiments on larger images, it becomes crucial to pre-compute anisotropic manifold graphs and their Laplacians. Dedicated librairies like PyKeops \citep{charlier2020kernel} enable this without memory issues. Nevertheless, the graph operations (convolutions, pooling or unpooling) still scale with the size of the graph. Fortunately, PyTorch provides sparse operations that increase efficiency in terms of time and memory compared to dense operations in cases of sufficiently sparse graph Laplacians (typically a sparsity $\mathcal{S}(\boldsymbol{\tilde{\Delta}}) \geq 98.5\%$). 

We evaluate our models on an image classification task on STL10 \citep{coates2011analysis} and an image segmentation task on ClimateNet \citep{kashinath2021climatenet}. We show the adaptability of our method by using a Wide Residual architecture \citep{zagoruyko2016wide} on STL10 and a U-Net-like network \citep{ronneberger2015u} on ClimateNet. We also demonstrate the potential of our approach and the benefits of using anisotropic spaces. Indeed, while on ClimateNet the use of anisotropies is neither beneficial nor detrimental, the difference in performance on STL10 is significant.

\begin{table}[h!]
\centering
\caption{Mean of test performance and training duration on ClimateNet and STL10. Errorbars are 1 standard deviation computed over 5 trials.}
\begin{tabular}{c c c c c c}
\toprule
 & & \multicolumn{2}{c}{ClimateNet} & \multicolumn{2}{c}{STL10} \\
$\epsilon$ & & Test F1 & Duration & Test accuracy & Duration \\
\midrule
$1$ & (\text{invariant}) & $\boldsymbol{85.62 \pm 0.09 \%}$ & $\sim \SI{2}{\day}$ & $68.98 \pm 0.56 \%$ & $\sim \SI{9}{\hour}$ \\
$0.1$ & (\text{equivariant}) & $85.25 \pm 0.19 \%$ & $\sim \SI{7}{\day}$ & $\boldsymbol{74.02 \pm 1.10 \%}$ & $\sim \SI{16}{\hour}$ \\
\bottomrule
\end{tabular}
\end{table}

\section{Conclusion} \label{sec:conclusion}


\paragraph{Scope.} With our method, geometric graph NNs are made equivariant to Lie groups. Via the groups $SE(2)$ and $SE(3)$, we can construct roto-translation equivariant networks for $2D$ image data and $3D$ volumetric data. Based on the group $SO(3)$, our method can deal with meteorological or cosmological data while preserving rotation equivariance. We believe that our flexible approach is ideal for further explorations on the relevance of group equivariance in tasks not considered in this work. 


\paragraph{Limitations.} The main weakness of our method is its relatively high memory requirement. Although all experiments ran on a single gpu, by adding an orientation axis, we significantly enlarge the feature maps. As a result, anisotropic graph manifolds are memory-heavier than isotropic ones and prone to a slowdown during the forward- and backward-pass. Nevertheless, with the emergence of geometric deep learning, we expect improvement in the hardware and implementation of graph-oriented operations. 
Another challenge is the increased number of hyper-parameters for which we only have derived rules of thumb. The graph connectivity and resolutions require a tradeoff between efficiency and quality of the manifold approximation. The anisotropic parameters require an analysis of the dataset and some intuition about the amount of anisotropy to set. With systematic hyper-parameter optimization, we can find an optimal combination, but requires more computational resources.

\paragraph{Potential and future research.} Thanks to its easy-to-tune anisotropic properties, our model can be used to better understand anisotropic properties in data. In particular, one could explore the effect of using anisotropic spaces instead of isotropic ones on many tasks and conclude when such anisotropic information is relevant. In this vein, it could also be interesting to derive anisotropic pooling and unpooling layers based on anisotropic spaces instead of isotropic ones as it is usually done. More generally, our method is simple enough to be extended to shapes/surfaces with a Riemannian manifold structure \citep{cohen2019gauge}, following the framework of coordinate independent convolutions \citep{weiler2021coordinate}. In this work, we focused on 2D images and spherical data on, but the method is readily extendable to higher dimensional Lie groups such as the $SE(3)$ group to obtain 3D roto-translation equivariant ChebLieNets. Moreover, our method for constructing anisotropic geometries could directly improve other successful Euclidean distance-based graph NNs such as \citep{satorras2021n} by making them fully equivariant. Last but not least, despite graph-based algorithms being computationally sub-optimal compared to CNNs, their flexibility is a real asset. We see high potential in the exploration of graph sparsification to reduce computational complexity. 


\begin{ack}
This paper is a collaborative effort between the LTS2 (EPFL) and the AMLab (UvA). Hence, we wish to express our deepest gratitude to both schools for this amazing opportunity. We also thank all the people working at SURFsara\footnote{\url{https://userinfo.surfsara.nl/}}, for letting us use Lisa, their excellent computation service.
\end{ack}

\clearpage
\bibliographystyle{plainnat}
\bibliography{bibliography}

\clearpage

\appendix

\section{Background} \label{app:background}

\subsection{Riemannian geometry} \label{app:riemannian_geometry}

\begin{figure*}[h!]
    \centering
    \includegraphics[width=\textwidth]{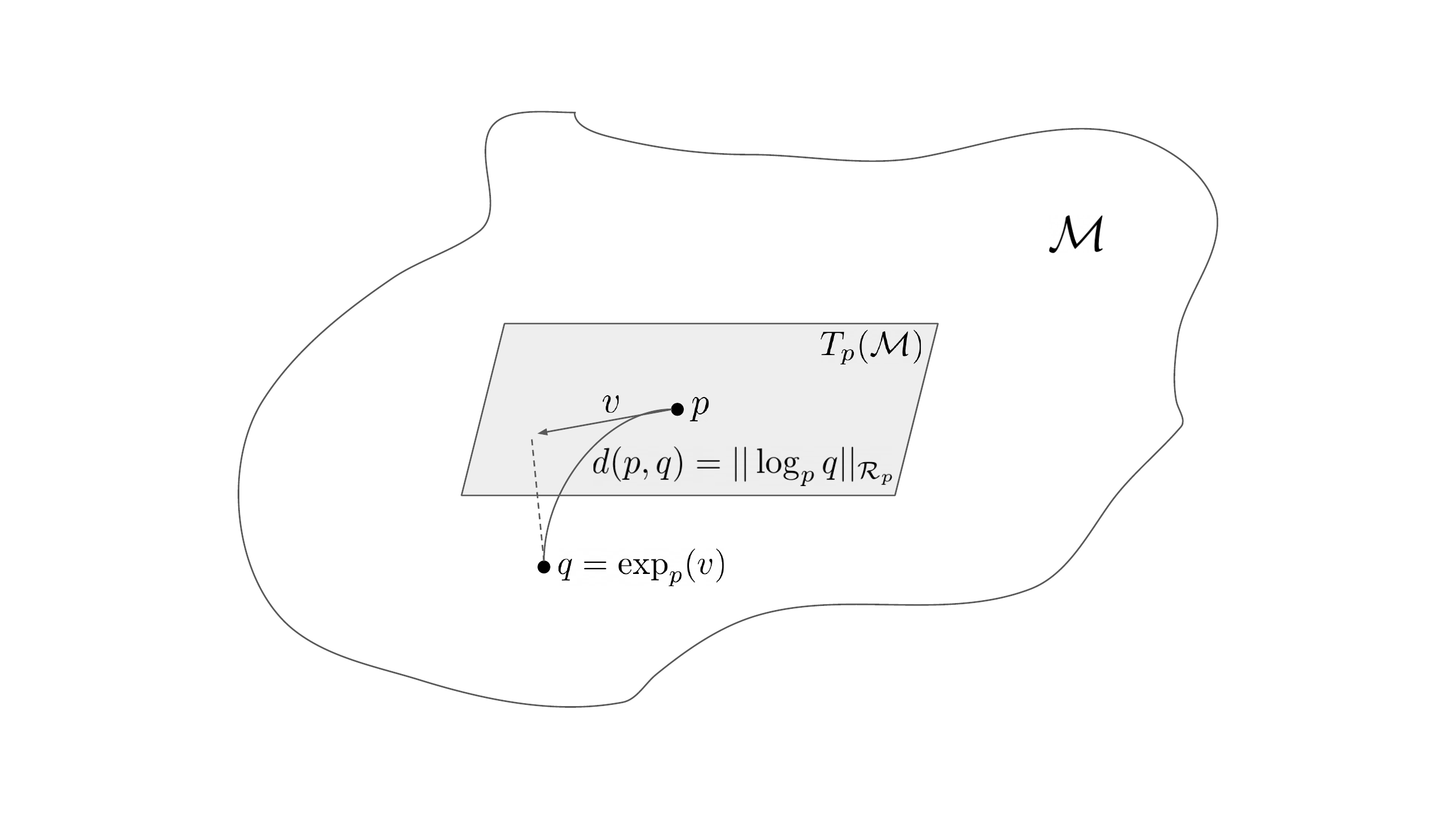}
    \caption{Riemannian exponential map $\exp_p : T_p(\mathcal{M}) \to \mathcal{M}$ and Riemannian logarithmic map $\log_p : \mathcal{M} \to T_p(\mathcal{M})$.}
    \label{fig:explogmap}
\end{figure*}

Riemannian geometry is a part of differential geometry that studies Riemannian manifolds, smooth manifolds equipped with a Riemannian metric. A manifold $\mathcal{M}$ is a generalization and abstraction of the notion of a curved surface. It is a topological set that is modeled closely on Euclidean space locally but may vary widely in global properties. It means for each $p \in \mathcal{M}$, one can associate a tangent space $T_p(\mathcal{M}) \subseteq \mathbb{R}^d$, corresponding to the union of all tangent vectors of differentiable curves passing through $p$. Consequently, at any point $p \in \mathcal{M}$, the tangent space $T_p(\mathcal{M})$ is spanned by a basis $\{\partial_{x_i, p} \}_{i=1}^N$:
\begin{equation}
T_p(\mathcal{M}) = \{ \dot{\gamma}(0) | \gamma : \mathbb{R} \to \mathcal{M} \in \mathcal{C}^1 \text{and } \gamma(0) = p\} = \spn\{\partial_{x_i,p} \}_{i=1}^N.
\end{equation}
A Riemannian manifold is a differentiable manifold $\mathcal{M}$ with a Riemannian metric $\mathcal{R}$, a 2-tensor field, such that at each $p \in \mathcal{M}$, we have a function $\mathcal{R}|_p : T_p (\mathcal{M}) \times T_p(\mathcal{M}) \to \mathbb{R}$ which is symmetric and positive definite. At each $g \in G$, the Riemannian metric can be expressed by its local representation, a symmetric positive definite matrix $\boldsymbol{R}_p$ whose components are given by:
\begin{equation}
r_{ij}(p) = \mathcal{R}|_p (\partial_{x_i, p}, \partial_{x_j, p}).
\end{equation}
In the following we consider Riemannian metric with  diagonal local representation at the origin $e$, that is:
\begin{equation}
r_{ij}(e) = \left\{
    \begin{array}{ll}
        r_{i} & \text{if } i = j  \\
        0 & \text{otherwise}
    \end{array}
\right..
\end{equation}
The Riemannian metric induces an inner product, such that for any $u = u^i \partial_{x_i, e}$ and $v = v^i \partial_{x_i, e}$, we have $\langle u, v \rangle_{\mathcal{R}(e)} = u^i r_i v^i$ and $|| u ||_{\mathcal{R}(e)}^2 = u^i r_i u^i$. Between each pair of points $p$ and $q$ of a Riemannian manifold, we define their Riemannian distance as the length of the shortest curve connecting the two points (a.k.a. geodesic). More formally, we have:
\begin{equation}
d(p, q) = \inf_{\gamma \in \mathcal{C}^\infty_{p, q}} \int_0^1 ||\dot{\gamma}(\tau)||_{\mathcal{R}_{\gamma(\tau)}} d\tau,
\end{equation}
where $\mathcal{C}^\infty_{p, q} = \{ \gamma : [0, 1] \to \mathcal{M} | \gamma \in \mathcal{C}^\infty, \gamma(0) = p,  \gamma(1) = q\}$. An more convenient way to define the Riemannian between two points is via the exponential and logarithmic Riemannian maps (Figure \ref{fig:explogmap}). The Riemannian exponential map $\exp_p T_p(\mathcal{M}) \to \mathcal{M}$ is defined as:
\begin{equation}
\exp_p (v) = \gamma_{p, v}(1) ,\qquad (\text{and by extension } \exp_p (tv) = \gamma_{p, v}(t)),
\end{equation}
where $\gamma_{p,v}(1)$ is the unique Riemannian geodesic starting at $p \in \mathcal{M}$ with initial velocity $v \in T_p(\mathcal{M})$. We can interpret the exponential map as follows. Let’s choose any direction in our tangent space and follow it with a step forward. We make sure to take the shortest path and end up at a new point. This process is what we called the exponential map.\footnote{It comes from the fact that all these tiny steps magically resemble the series expansion of the exponential function.} The inverse mapping of the exponential map, the logarithmic map $\log_p : \mathcal{M} \to T_p(\mathcal{M})$ will map an element $p \in \mathcal{M}$ to the smallest vector $v = \log_p q \in T_p(\mathcal{M})$ as measured by the Riemannian metric such that $q = \exp_p v \in \mathcal{M}$.

\subsection{Group theory} \label{app:group_theory}

A group $(G, \cdot)$ is a set $G$ equipped with a binary operation $\cdot : G \to G$ called group product, satisfying the four group axioms (closure, associativity, identity element and inverse elements). To map the structure of the group to some mathematical object, one requires a representation. We define $H$ as the vector space to which our mathematical object belongs and $\mathcal{B}(H)$ the space of bounded linear invertible operators $H \to H$. A representation $\mathcal{V} : G \to \mathcal{B}(H)$ maps a group element to an operator such that the identity element, the group product and the group inverse are preserved. We define the left-regular representation $\mathcal{L}_g$ on the (infinite-dimensional) vector space of functions $G \to \mathbb{R}^d$ via:
\begin{equation}
(\mathcal{L}_g \circ f )(h) = f(g^{-1}h),
\end{equation}
with $f : G \to \mathbb{R}^d$ a function on the group $G$ and $g, h$ elements of the group $G$. Using this group representation, we can formally define the left-invariance (a.k.a. equivariance).

\begin{definition}[Equivariance] \label{def:equivariance}
An operator $\Phi : \mathcal{X} \to \mathcal{Y}$ from one vector space to the other is equivariant (or left-invariant under group transformation) if it satisfies the following property:
\begin{equation}
\mathcal{L}'_g \circ \Phi = \Phi \circ \mathcal{L}_g, \qquad \forall g \in G.
\end{equation}
\end{definition}

Equivariance can be realized in many ways, and in particular, the group representations $\mathcal{L}_g$ and $\mathcal{L}'_g$ need not be the same, as they act on different spaces $\mathcal{X}$ and $\mathcal{Y}$. Note that the familiar concept of invariance is a special kind of equivariance where  $\mathcal{L}_g'$ is the identity transformation for all element $g$ of the group $G$.

A Lie group is a continuous group whose group elements are parameterized by a finite-dimensional differentiable manifold. In essence, this means that a Lie group is a group to which we can apply differential geometry. From now on, we also assume the group manifold is equipped with a Riemannian metric, that is it is Riemannian manifold. To each element $g$ of $G$, we can associate a tangent space $T_g(G)$, which is spanned by a basis of left-invariant vectors denoted by $\{\mathcal{A}_i|_g\}_{i=1}^n$. We write $T_g(G) = \spn\{\mathcal{A}_1|_g, \dots, \mathcal{A}_n|_g\}$ and we can express tangent vectors $\dot{\gamma}(t)$ of $T_g(G)$ in this basis via $\dot{\gamma}(t) = c^i(t) \mathcal{A}_i|_{\gamma(t)}$.

Moreover, because this frame of basis vector is left-invariant, the coefficients $c_i(t)$ remain unchanged if the curve is moved by applying left group product. The tangent space at the origin $T_e(G)$ is spanned by a basis $\{A_i\}_{i=1}^n$ where $A_i = \mathcal{A}_i|_e$. We make a subtle difference in notation between $\mathcal{A}$ and $A$, where $\mathcal{A}$ represents a whole vector field and $\mathcal{A}|_g$ represents the vector at location $g$. The straight $A$ is used to indicate a vector in the Lie algebra, the tangent space at the origin. Via the push-forward $(L_g)_*$, we can generate a whole vector space by picking a vector in the Lie algebra and transporting tangent vectors from $\dot{\gamma}(0) \in T_e(G)$ to $g \cdot \dot{\gamma}(0) \in T_g(G)$ using:
\begin{equation}
\mathcal{A}_i|_g = (L_g)_* A_i , \qquad \forall g \in G.
\end{equation}

The left-invariant frame of basis vectors could also be interpreted as the directional derivative of functions defined on the group $G$. At any element $g$, the value of the directional derivative of a function $f$ defined on $G$ can be computed using the push-forward operation:
\begin{equation}
\mathcal{A}_i|_g f = (L_g)_* \mathcal{A}_i|_e f := \mathcal{A}_i|_e (f \circ L_{g^{-1}}),
\end{equation}
so $\mathcal{A}_i|g f$ represents the directional derivative along the vector field at location $g$, which can be defined by translating the function back to the origin via $L_{g^{-1}}$ and compute the derivative at the origin along the direction specified by $\mathcal{A}_i|_e$. In the above $L_g h:= g h$.

The Laplace-Beltrami operator, the generalization to Riemannian manifold of the Laplace operator\footnote{Another equivalent definition in local coordinates exists too. It explicitly takes into account the Riemannian metric tensor field $\mathcal{R}$ previously defined. Here we use a more compact definition but using the subscript $\mathcal{R}(g)$ to indicate that the Laplace-Beltrami operator depends on the location on the Riemannian manifold and the Riemannian metric tensor field.} is defined below.
\begin{definition}[Laplace-Beltrami operator] \label{def:laplace_beltrami}
Let $g$ be an element of the Lie group $(\mathcal{M}, \cdot)$. The Laplace-Beltrami operator on the Riemannian manifold $\mathcal{M}$ is defined as
\begin{equation}
\Delta_{\mathcal{R}(g)} = \dive (\nabla_{\mathcal{R}(g)})
\end{equation}
\end{definition}

Using the left-invariant frame of basis vectors as directional differential operators, we can define the gradient, expressed as a vector relative to the basis $\mathcal{A}_i|_g$, of a function $f : G \to G$:
\begin{equation}
\nabla_{\mathcal{R}(g)} f = \mathbf{R}_{g}^{-1} (\mathcal{A}_1|_g f, \dots, \mathcal{A}_n|_g f)^\top,
\end{equation}
where $\mathbf{R}_{g}$ is the Riemannian metric tensor defined relative to the basis $\mathcal{A}_i$, that corrects for local scaling and shrinking of the manifold as measured by the metric tensor.
The divergence of a vector field $F : G \to \mathbb{R}^n$ is the operator:
\begin{equation}
\dive(F) = \sum_{i=1}^n \mathcal{A}_i|_g F_i.
\end{equation}

The Laplace-Beltrami operator depends on a Riemannian metric tensor, which describes how lengths of vectors should be measured in different directions, and in the Laplace-Beltrami operator, it rescales the derivates accordingly. While the usual Laplacian is isotropic (derivatives are threated the same in each direction), the Laplace-Beltrami operator can be anisotropic due to the Riemannian metric that is used.

\begin{theorem}[Left-invariance of the Laplace-Beltrami operator] \label{thm:equiv_laplace_beltrami}
The Laplace-Beltrami operator $\Delta_{\mathcal{R}(g)}$ is left-invariant and satisfies:
\begin{equation}
\Delta_{\mathcal{R}(g)} = (L_g)_* \Delta_{\mathcal{R}(e)}.
\end{equation}
\end{theorem}

A Lie algebra $\mathfrak{g}$ is a vector space (here the tangent space at the identity element $T_e(G)$) that is endowed with a binary operator called the Lie bracket or commutator $[\cdot, \cdot] : T_e(G) \times T_e(G) \to T_e(G)$ that is bilinear, alternative and satisfies the Jacobi identity. Conceptually, the Lie bracket $[A, B]$ of two vector fields $A, B$ is the derivative of $B$ along the flow generated by $A$.\footnote{Two vector fields are commutative (zero Lie bracket) if and only if its flows are too, in the sense that there is no difference starting at one point $p$, traveling a time $t_a$ over the flow of $A$ and then a time $t_b$ over the flow of $B$, or, instead, traveling first $t_b$ over the flow of $B$ and then $t_a$ over the flow of $A$.} In the following, we take $\{A_i\}_{i=1}^n$ with $A_i = \mathcal{A}_i|_e$ as Lie Algebra and get left-invariant vector fields via the push-forward operation.

The Lie group exponential and logarithmic maps define the mappings between the group and the tangent space. The exponential map on a Lie group can be thought of as picking a vector $A$ in the Lie algebra, construct a left-invariant vector field $\mathcal{A}$ via the push-forward operator, and follow this vector field by taking infinitesimal steps along the direction indicated by the vector field. Integrating along the vector field defined by $A$ for unit time brings to some point $g \in G$. So $g = \exp A$, where $\exp: \mathfrak{g} \to G$.\footnote{Note that Riemannian and Lie group exponential maps are different. For the general Riemannian exponential map, the vector field along which we compute the path integral is defined by the Riemannian metric. In the Lie group exponential map, the vector field is defined by the push forward of left-multiplication. The curves defined by the path integrals in the Riemannian exponential are known to be geodesics. The exponential curves in the Lie group case are "straight curves" with respect to a moving frame of reference but are not necessarily geodesics. The Riemannian and Lie group exponential maps only coincide when the Riemannian metric is both left and right invariant (see Section 4.5 in \citet{bekkers2017thesis}).}

\subsection{Graph theory} \label{app:graph_theory}

Graphs are generic data representation forms that are useful for describing the geometric structure of data domains \citep{west1996introduction}. More formally, we define a graph $\mathcal{G}$ as a structure modeling a finite set of interactions called edges $\mathcal{E}$ between a finite set of objects called vertices $\mathcal{V}$. In the following, we denote by $|\mathcal{V}|$ the number of vertices and by $|\mathcal{E}|$ the number of edges. We denote $v_i$ the $i$-th vertex and $e(v_i, v_j)$ the potential edge from $v_i$ to $v_j$. We call the neighborhood of the vertex $v_i$ the set of vertices connected to $v_i$ by an edge and denote it by $\mathcal{N}(v_i)$. More generally, we write $\mathcal{N}^k(v_i)$ for the $k$-hops neighborhood of vertex $v_i$, that is, the set of vertices connected to $v_i$ with a path of at most $k$ edges. In some cases, it can be useful to add weights on graph edges. In general, the weights can take any value. Nevertheless, in this thesis, we assume weights in the range $[0, 1)$ and measuring the similarity rather than the distance. When the edges' weights are not naturally defined by an application, a common way to define them is to apply a kernel $\mathfrak{K} : \mathbb{R}_+ \to [0, 1)$ on the distance between connected vertices. For theoretical convergence results, we use a Gaussian weighting scheme:
\begin{equation}
w(v_i, v_j) =
\left\{
\begin{array}{ll}
\exp \left(- \frac{d^2(v_i, v_j)}{4t}\right) & \text{if } e(v_i, v_j) \in \mathcal{E} \\
0 & \text{otherwise}
\end{array}
\right.,
\end{equation}
where $d(v_i, v_j)$ denotes the (Riemannian) distance between vertices $v_i$ and $v_j$ and $t$ is a positive real number called bandwidth of the Gaussian kernel. From now on, we furthermore assume that the graphs are undirected and without self-loop.

The field of signal processing on graphs merges algebraic and spectral graph theoretic concepts with computational harmonic analysis to process such signals on graphs \citep{shuman2013gsp}. A signal on a graph is a function $f : \mathcal{V} \to \mathbb{R}^d$, mapping each vertex of the graph to a $d$-dimensional real valued vector. In matrix form, this signal is a $|\mathcal{V}| \times d$ real valued matrix $\boldsymbol{f}$ whose rows are given by $\boldsymbol{f}_i = f(v_i) \in \mathbb{R}^d$.

An essential object in graph signal processing is the Laplacian operator. Under some specific conditions that we will state later, it can be interpreted as a discrete version of the Laplace-Beltrami operator.
\begin{definition}[Symmetric normalized Laplacian] \label{def:graph_laplacian}
Let $\mathcal{G}$ be an undirected weighted graph without self-loops. The symmetric normalized Laplacian $\boldsymbol{\Delta}$ is the $|\mathcal{V}| \times |\mathcal{V}|$ real valued matrix whose components are given by:
\begin{equation}
\Delta_{i, j} =
\left\{
\begin{array}{ll}
1 & \text{if } i = j \text{ and } \deg(v_i) > 0 \\
- \frac{w(v_i, v_j)}{\sqrt{\deg (v_i) \deg (v_j)}} & \text{if } i \neq j \text{ and } \deg(v_i) > 0 \\
0 & \text{otherwise}
\end{array}
\right..
\end{equation}
\end{definition}

Assuming a function $f$ defined on the graph vertices $\mathcal{V}$, by inspection on the components of $\boldsymbol{\Delta}^k \boldsymbol{f}$, we remark that the Laplacian acts as a $k$-op neighborhood operator. We can prove that $\boldsymbol{\Delta}$ is a symmetric positive definite matrix. Hence it admits a unique eigendecomposition of the form $\boldsymbol{\Delta} = \boldsymbol{\Phi \Lambda \Phi}^\top$ where the $j$-th column of $\Phi$ correponds to the eigenvector $\phi_j$ associated with real positive eigenvalue $\lambda_j$. In these settings, the eigenvalues are in the range $[0, 2]$ \citep{chung1997spectral}. By analogy with the Euclidean case where the Laplacian's eigenfunctions correspond to the Fourier basis, we can construct a graph Fourier basis from the eigendecomposition of the graph Laplacian, and define the graph Fourier transform and its inverse.\footnote{The existence of an inverse transform is a direct consequence of the orthonormality of the eigenvectors.}
\begin{definition}[Graph Fourier transform]
Let $\mathcal{G} = (\mathcal{V}, \mathcal{E})$ be a graph with Laplacian $\boldsymbol{\Delta}$ and let $f : \mathcal{V} \to \mathbb{R}$ be a signal defined on the graph's vertices. The graph Fourier transform $\hat{f}$ of $f$ is given by:
\begin{equation}
\hat{f}(\lambda_j) = \hat{f}_j = \{ \boldsymbol{\Phi}^\top \boldsymbol{f}\}_j = \sum_i \phi_{ij} f_i = \sum_{v_i} \phi(v_i, \lambda_j) f(v_i),
\end{equation}
and its inverse transform by:
\begin{equation}
f(v_i) = f_i = \{\boldsymbol{\Phi}  \boldsymbol{\hat{f}} \}_i = \sum_j \phi_{ij} \hat{f}_j = \sum_{\lambda_j} \phi(v_i, \lambda_j) \hat{f}(\lambda_j).
\end{equation}
\end{definition}

\section{Lie groups} \label{app:lie_groups}

At the moment, all the group's elements we are interested in are in 1-1 correspondence with elements of the 3-dimensional general linear group $GL(3)$, the group of $3 \times 3$ real matrices. We will use this representation since working with matrices is more convenient. Indeed, in this case the group product, group inverse, group exponential and group logarithm respectively coincide with the matrix product, matrix inverse, matrix exponential and matrix logarithm.

For the roto-translation group  $SE(2)$, the spatial part corresponds to the planar translations and the orientation part to the rotation angles. Group elements $g \in SE(2)$ are given in matrix formulation by:
\begin{equation}
g = (x, y, \theta) \leftrightarrow
\boldsymbol{G_g} =
\left(
\begin{array}{ccc}
\cos \theta & - \sin \theta & x \\
\sin \theta & \cos \theta & y \\
0 & 0 & 1 \\
\end{array}
\right),
\end{equation}
hence the group is 3 dimensional with two parameters $x,y$ coming from the spatial space $\mathbb{R}^2$ and one orientation parameter coming from $[-\pi, \pi)$.

The Lie algebra $\mathfrak{se}(2)$ is the set of $3 \times 3$ matrices:
\begin{equation}
\boldsymbol{A_1} =
\left(
\begin{array}{ccc}
0 & 0 & 1 \\
0 & 0 & 0 \\
0 & 0 & 0 \\
\end{array}
\right)
, \quad
\boldsymbol{A_2} =
\left(
\begin{array}{ccc}
0 & 0 & 0  \\
0 & 0 & 1 \\
0 & 0 & 0 \\
\end{array}
\right)
, \text{ and} \quad
\boldsymbol{A_3} =
\left(
\begin{array}{ccc}
0 & -1 & 0 \\
1 & 0 & 0 \\
0 & 0 & 0 \\
\end{array}
\right)
\end{equation}

Naturally, it is impossible to uniformly sample elements on this group because the $\mathbb{R}^2$ space is infinite. Adding boundaries to the euclidean space, we get the $[0, 1)^2$ space, which is not homogeneous anymore, but on which it is possible to uniformly sample $|\mathcal{V}_s|$ elements. Because we are considering anisotropic Laplace-Beltrami operators which are symmetric under reflections, it is sufficient to sample the rotation angles $\theta$ in the range $[-\pi/2, \pi/2)$.

As defined by its Lie algebra, the matrix logarithm related to an element of the $SE(2)$ group has the form:

\begin{equation}
\log_e \boldsymbol{G}_g =
\left(
\begin{array}{ccc}
0 & -c_3 & c_1 \\
c_3 & 0 & c_2 \\
0 & 0 & 0 \\
\end{array}
\right)
\end{equation}

It results that the logarithmic map $\log_e : SE(2) \to \mathfrak{se}(2)$ admits a closed form expression $\log_e g = (c_1, c_2, c_3)^\top$ with\footnote{In the isotropic case, corresponding to the $2$-d Euclidean space, one can check that using the exact same expression of the logarithmic map with $\theta = 0$ gives the Euclidean distance.}:
\begin{equation}
c_1 = \frac{\theta}{2} \left(y + x \cot \frac{\theta}{2} \right)
, \quad
c_2 = \frac{\theta}{2} \left(-x + y \cot \frac{\theta}{2} \right)
, \text{ and} \quad
c_3 = \theta.
\end{equation}

The group $SO(3)$ of all rotations about the origin of 3-dimensional Euclidean space can be split into a "spatial" part which is the sphere, and a rotation part, which is a rotation around a particular reference axis. Using $ZYZ$ representation, group elements $g \in SO(3)$ are given in matrix formulation by:
\begin{equation}
g = (\alpha, \beta, \gamma) \leftrightarrow
\boldsymbol{G_g} = \boldsymbol{R}_{\gamma, z} \boldsymbol{R}_{\beta, y} \boldsymbol{R}_{\alpha, z},
\end{equation}
where $\alpha \in [-\pi, \pi]$, $\beta \in [-\pi/2, \pi/2]$ and $\gamma \in [-\pi, \pi]$. We view the sphere $S^2$ as the spatial part, just like we view our Earth as locally flat. $S^2$ is parametrized with Euler angles $\beta$ and $\gamma$, which are independent of $\alpha$. The rotation part is then parametrized by $\alpha$.

The Lie algebra $\mathfrak{so}(3)$ is the set of antisymmetric $3 \times 3$ matrices:
\begin{equation}
\boldsymbol{A_1} =
\left(
\begin{array}{ccc}
0 & 0 & 1 \\
0 & 0 & 0 \\
-1 & 0 & 0 \\
\end{array}
\right)
,\quad
\boldsymbol{A_2} =
\left(
\begin{array}{ccc}
0 & -1 & 0  \\
1 & 0 & 0 \\
0 & 0 & 0 \\
\end{array}
\right)
, \text{ and} \quad
\boldsymbol{A_3} =
\left(
\begin{array}{ccc}
0 & 0 & 0 \\
0 & 0 & -1 \\
0 & 1 & 0 \\
\end{array}
\right).
\end{equation}

To uniformly sample on the group $SO(3)$ is quite challenging because a perfectly uniform sampling on the sphere does not exist. Fortunately, this task has been largely studied, and many algorithms have been proposed: equiangular\footnote{Equiangular is far from uniform, but it has a sampling theorem.} \citep{driscoll1994computing}, HEALPix \citep{gorski2005healpix}, and icosahedral \citep{baumgardner1985icosahedral} samplings. As before, due to the same symmetry argument, it is sufficient to restrict the orientation range to $[-\pi/2, \pi/2)$ and to sample $|\mathcal{V}_o|$ elements on this restricted range.

The matrix logarithm related to an element of the $SO(3)$ group has the form:

\begin{equation}
\log_e \boldsymbol{G}_g =
\left(
\begin{array}{ccc}
0 & -c_3 & c_2 \\
c_3 & 0 & -c_1 \\
-c_2 & c_1 & 0 \\
\end{array}
\right).
\end{equation}

Using the Rodrigues' rotation formula \citep{rodrigues1840lois}, we derive a closed form expression for the logarithmic map $\log_{e} : SO(3) \to \mathfrak{so}(3)$, with notation $\log_{e} g = (c_1, c_2, c_3)^\top$. We get\footnote{To compute the logarithmic map in the isotropic case corresponding to $S^2$, we must use a slightly modified version such that $c_3 = 0$, which via the exponential map generate torsion free exponential curves. The logarithmic of any rotation matrix defined by $(-\gamma, \beta, \gamma)$ yields this results \citep{bekkers2019b, portegies2015new}.}:

\begin{equation}
c_1 = \frac{\theta}{2 \sin \theta} ( \boldsymbol{G}_{g, 3, 2} - \boldsymbol{G}_{g, 2, 3})
, \quad
c_2 = \frac{\theta}{2 \sin \theta} ( \boldsymbol{G}_{g, 2, 1} - \boldsymbol{G}_{g, 1, 2})
, \text{ and} \quad
c_3 = \frac{\theta}{2 \sin \theta} ( \boldsymbol{G}_{g, 1, 3} - \boldsymbol{G}_{g, 3, 1}).
\end{equation}

Due to the $\pi$ periodicity of the orientation axis, the $\pi$-periodic Riemannian distance between two elements of the groups is the minimal distance between:
\begin{itemize}
\item the original distance without offset;
\item the original distance with a negative offset $-\pi$ on the orientation axis;
\item the original distance with a positive offset $+\pi$ on the orientation axis.
\end{itemize}

\section{Experiment details} \label{app:experiment_details}

Empirical evidence showed that deep and wide neural networks are keys to good performances. To help going deeper we use residual convolutional layers \citep{he2016deep} and batch normalization \citep{ioffe2015batch}. To avoid tuning of the learning rate, we use ADAM optimizers \citep{kingma2014adam} for all our experiments. Last, we initialise our models' parameters via the Kaiming method \citep{he2015delving}.

\paragraph{Stability test on MNIST.} In these experiments, we used a Wide Residual ChebNet with three convolutional residual layers with rectified linear units and kernel of size $4$. As we do not pool at all, we only define one $SE(2)$ 16-NN graph with $28 \times 28 \times 6 = 4704$ vertices. We use extreme spatial anisotropy with $\epsilon^2 = 0.1$ and reach the 40/60 ratio by setting $\xi$ accordingly. The residual layers are followed by a projective global max pooling layer and a fully connected with LogSoftMax output layer.

\paragraph{Orientation anisotropic test on CIFAR10.} In these experiments, we used a Wide Residual ChebNet with three residual layers with rectified linear units and kernel of size $4$. We use two R2RandPool layers. Hence we define three $SE(2)$ 16-NN graphs with $32 \times 32 \times 6= 6144$, $16 \times 16 \times 6 = 1536$, and $8 \times 8 \times 6 = 384$ vertices. We use extreme spatial anisotropy with $\epsilon^2 = 0.1$. The residual layers are followed by a projective global max pooling layer and a fully connected with LogSoftMax output layer.

\paragraph{Spatial anisotropic test on CIFAR10.} In these experiments, we used a Wide Residual ChebNet with 3 residual layers with rectified linear units and kernel of size $4$. We use two R2RandPool layers. Hence we define three $SE(2)$ 16-NN graphs with $32 \times 32 \times 6= 6144$, $16 \times 16 \times 6 = 1536$, and $8 \times 8 \times 6 = 384$ vertices. We use moderate orientation anisotropies to reach the 40/60 ratio on each graphs by setting $\xi$ accordingly. The isotropic case is constructed using $\epsilon^2 = \xi^2 = 1$ on three $\mathbb{R}^2$ 8-NN graphs with $32 \times 32 \times 1 = 784$, $16 \times 16 \times 1 = 256$, and $8 \times 8 \times 1 = 64$ vertices. The residual layers are followed by a projective global max pooling layer and a fully connected with LogSoftMax output layer.

\paragraph{Scalability test on STL10.} In these experiments, we used a Wide Residual ChebNet with three residual layers with rectified linear units and kernel of size $4$. We use two R2RandPool layers. Hence we define three $SE(2)$ 16-NN graphs with $96 \times 96 \times 6= 55296$, $48 \times 48 \times 6 = 13824$, and $8 \times 8 \times 6 = 3456$ vertices. We use extreme spatial anisotropy with $\epsilon^2 = 0.1$ and reach the 40/60 ratio on each graphs by setting $\xi$ accordingly. The isotropic case is constructed using $\epsilon = \xi = 1$ on three $\mathbb{R}^2$ 8-NN graphs with $96 \times 96 \times 1 = 9216$, $48 \times 48 \times 1 = 2304$, and $24 \times 24 \times 1 = 576$ vertices. The residual layers are followed by a projective global max pooling layer and a fully connected with LogSoftMax output layer. The residual layers are followed by a projective global max pooling layer and a fully connected with LogSoftMax output layer.

\paragraph{Scalability test on ClimateNet.} In these experiments, we used a U-ChebNet with three residual layers with rectified linear units and kernel of size $3$. We use five S2MaxPool (encoding) and five S2AvgUnpool (decoding) layers. Hence we define six $SO(3)$ 16-NN graphs with $10242 \times 6= 61452$, $2562 \times 6 = 15372$, $642 \times 6 = 3852$, $162 \times 6 = 972$, $42 \times 6 = 252$, and $12 \times 6 = 72$ vertices. We use extreme spatial anisotropy with $\epsilon^2 = 0.1$ and reach the 40/60 ratio on each graphs by setting $\xi$ accordingly. The residual layers are followed by a projective global max pooling layer and a fully connected with LogSoftMax output layer.

\section{Stability under random perturbations}

\begin{figure*}[h!]
    \centering
    \begin{subfigure}[t]{0.48\textwidth}
        \centering
        \includegraphics[width=\textwidth]{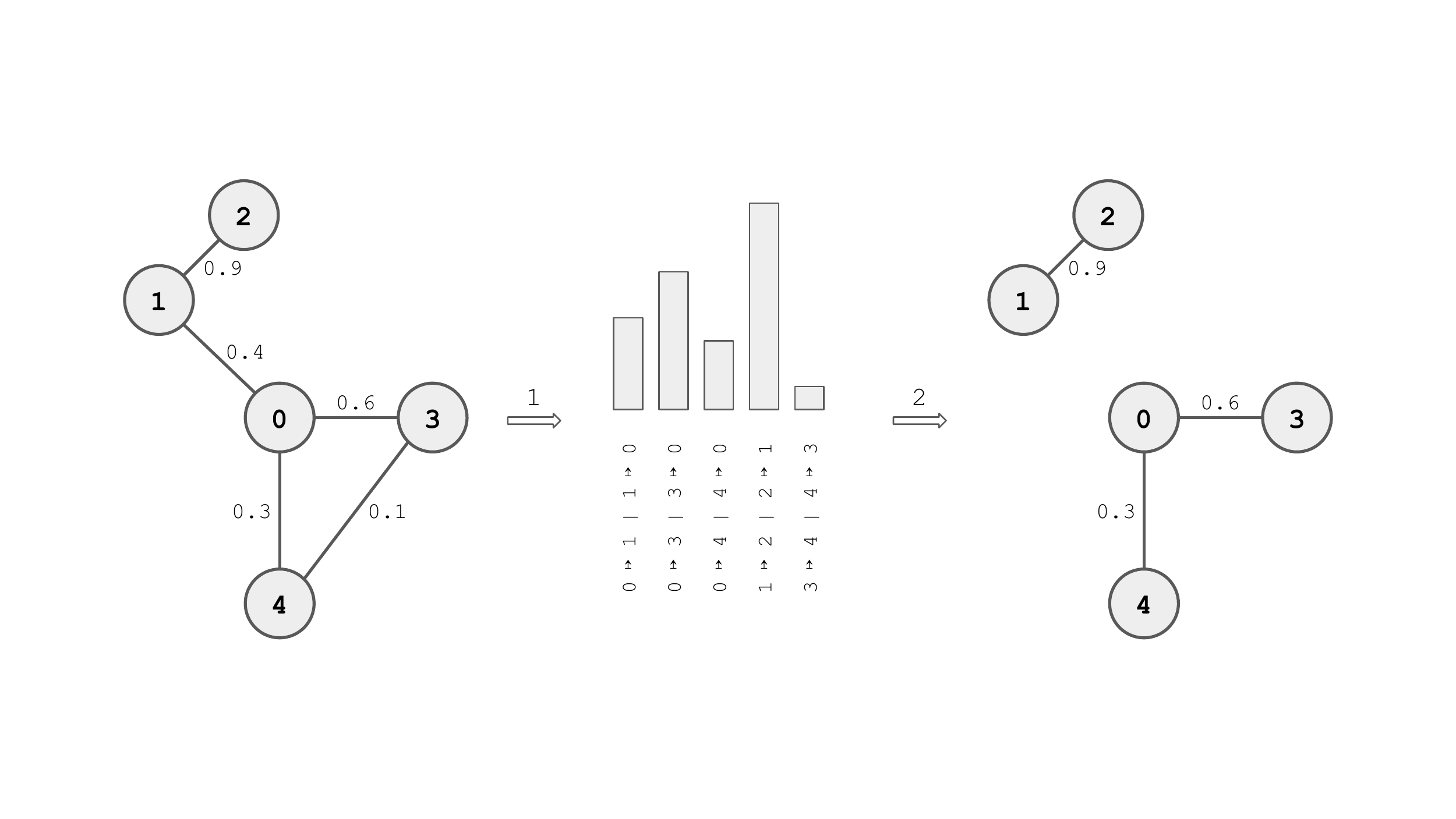}
        \caption{One constructs a random sub-graph by randomly sampling a rate~$\kappa_\mathcal{E}$ of edges of the complete graph based on their weight. As a direct consequence of the work of  \cite{keriven2020convergence}, the Laplacian of a quasi-sparse graph with randomly sampled edges remains consistent. Then, the unsampled edges are pruned by setting their weight to zero. One must be aware that this process could lead to isolated vertices or cluster of vertices. It is not necessarily a problem, but it is important to be mindful of this potential effect.}
    \end{subfigure}
    \hfill
    \begin{subfigure}[t]{0.48\textwidth}
        \centering
        \includegraphics[width=\textwidth]{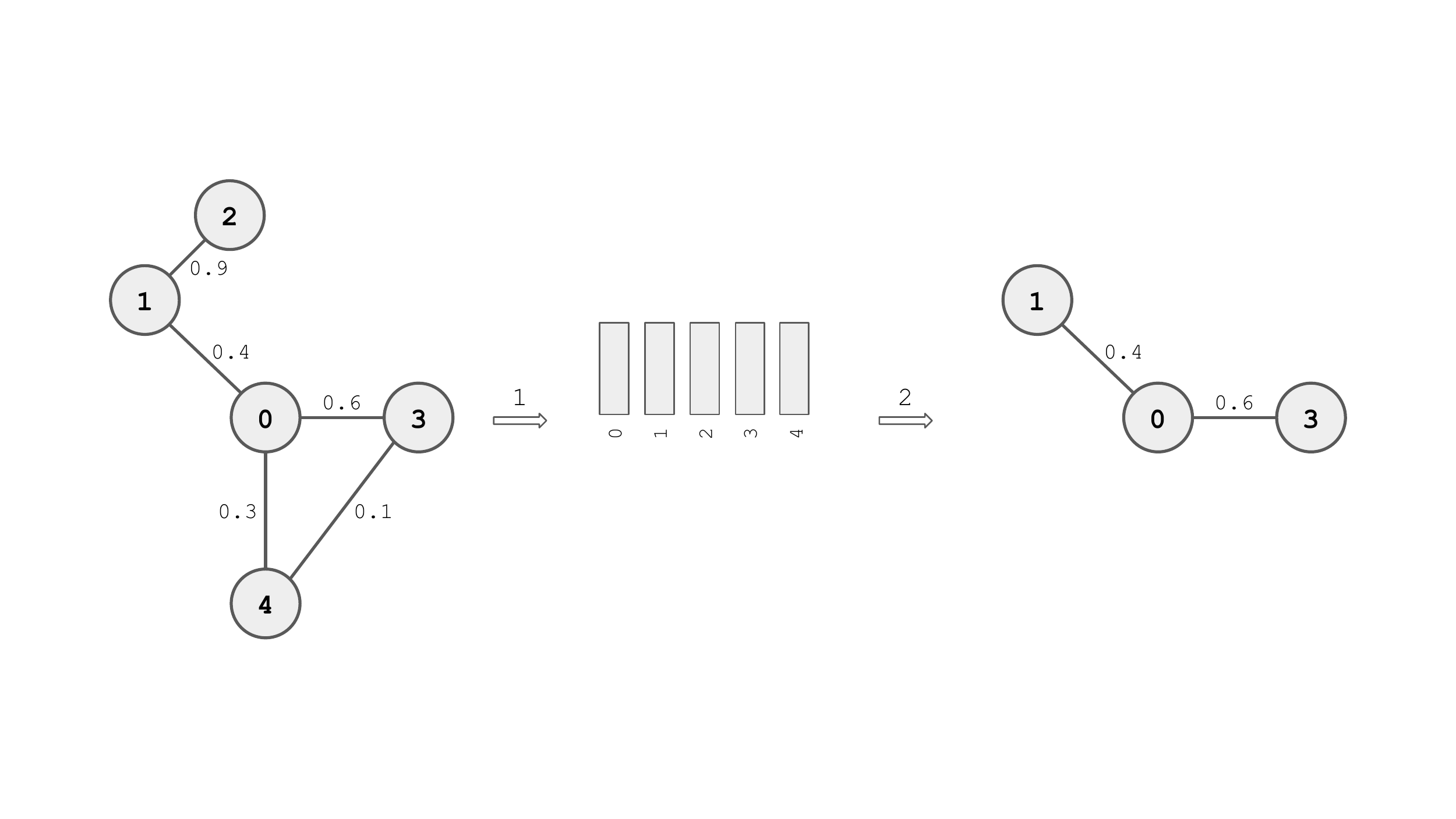}
        \caption{This method based on vertices sampling consists of creating a random sub-graph by randomly sampling a rate $\kappa_\mathcal{V}$ of vertices from the complete graph. The edges between one or mode unsampled vertices are discarded too. This method is more drastic and should be used carefully, as such a vertices pruning leads to a violation of the uniform distribution of vertices.}
    \end{subfigure}
    \caption{Edge- and vertex-based sampling methods.}
    \label{fig:sampling}
\end{figure*}

In sensitive domains, the stability of a neural network to random perturbation is a desired property. We aim at demonstrate the high stability of our approach, as random perturbations are added to graphs during the training. We introduce two methods, both consisting in randomly pruning the original graph to construct a random sub-graph (see Figure~\ref{fig:sampling}). At test time, random perturbations are removed in order to evaluate a perturbation-free model.

We run a bunch of experiments with a Wide Residual architecture on MNIST \citep{lecun-mnisthandwrittendigit-2010}, varying the rates of edges or vertices to sample. The objective of these experiments is twofold. First, we would like to test the stability of the model under random perturbations. Second, this experiment is also a good manner to demonstrate the equivariance property of our method. In this purpose, we train our model on the original training set of MNIST, adding some perturbations. At test time, we evaluate the perturbation-free model on the test set with and without random rotations.

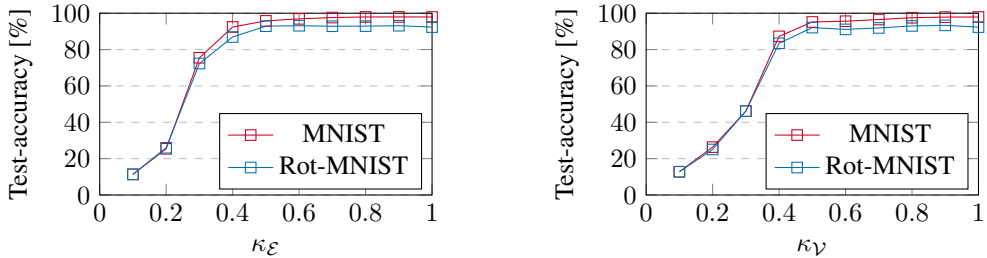
\begin{figure*}[h!]
    \centering
    \begin{subfigure}[b]{0.48\textwidth}
        \centering
        \begin{tikzpicture}
        \begin{axis}[
            xlabel={$\kappa_{\mathcal{E}}$},
            ylabel={Test-accuracy [\%]},
            xmin=0, xmax=1,
            ymin=0, ymax=100,
            xtick={0,0.2,0.4,0.6,0.8,1.0},
            ytick={0,20,40,60,80,100},
            legend pos=south east,
            ymajorgrids=true,
            grid style=dashed,
            width=6cm,
            height=4cm
        ]

        \addplot[
            color=color1,
            mark=square,
            ]
            coordinates {
            (0.1,11.35)(0.2,25.42)(0.3,75.49)(0.4, 92.43)(0.5,95.79)(0.6,96.89)(0.7,97.61)(0.8,97.94)(0.9,97.96)(1.0,97.92)
            };
            \addlegendentry{MNIST}

        \addplot[
            color=color2,
            mark=square,
            ]
            coordinates {
            (0.1,11.35)(0.2,26.00)(0.3,72.33)(0.4, 86.91)(0.5,92.87)(0.6,93.12)(0.7,92.84)(0.8,92.90)(0.9,93.12)(1.0,92.37)
            };
            \addlegendentry{Rot-MNIST}
        \end{axis}
        \end{tikzpicture}
        \caption{Test-accuracy against rate of edges to sample}
    \end{subfigure}
    \hfill
    \begin{subfigure}[b]{0.48\textwidth}
        \centering
        \begin{tikzpicture}
        \begin{axis}[
            xlabel={$\kappa_{\mathcal{V}}$},
            ylabel={Test-accuracy [\%]},
            xmin=0, xmax=1,
            ymin=0, ymax=100,
            xtick={0,0.2,0.4,0.6,0.8,1.0},
            ytick={0,20,40,60,80,100},
            legend pos=south east,
            ymajorgrids=true,
            grid style=dashed,
            width=6cm,
            height=4cm
        ]

        \addplot[
            color=color1,
            mark=square,
            ]
            coordinates {
            (0.1,12.75)(0.2,26.34)(0.3,46.15)(0.4, 87.27)(0.5,95.18)(0.6,95.59)(0.7,96.53)(0.8,97.52)(0.9,97.87)(1.0,97.92)
            };
            \addlegendentry{MNIST}

        \addplot[
            color=color2,
            mark=square,
            ]
            coordinates {
            (0.1,12.81)(0.2,25.04)(0.3,46.15)(0.4, 83.48)(0.5,92.09)(0.6,91.1)(0.7,91.9)(0.8,92.99)(0.9,93.32)(1.0,92.37)
            };
            \addlegendentry{Rot-MNIST}
        \end{axis}
        \end{tikzpicture}
        \caption{Test-accuracy against rate of vertices to sample}
    \end{subfigure}
    \caption{Stability and group equivariance under random perturbations.}
\end{figure*}

\paragraph{Equivariance.} With these experiments, we empirically demonstrated the rotation-equivariance of a ChebLieNet. An extrinsic equivariance error can explain the slight difference in performance between the original test set and the randomly rotated one. This error is not related to the model but the evaluation's method and can be semantic (e.g. is it always possible to differentiate a six from a nine without orientation information) or numerical (e.g., how to rotate a low-resolution image by 10 degrees without alteration). 

\paragraph{Robustness.} While in this specific case, our dropout-like method does not help to improve our model on the test set, we expect it could help to reduce the over-fitting on other tasks. In addition, one could notice that even at middle-low sampling rates, the models remain stable. First of all, it supports the idea that there is no point using a fully connected graph; more edges capture the geometry better but with diminishing returns. Secondly, the model is robustly able to catch the semantic information in an image, even with partial alterations.

\section{Empirical convergence in eigenmaps of the graph Laplacians} \label{app:laplacian}

As it has been proven by \cite{belkin2006convergence}, assuming a suitably constructed graph, the graph Laplacian converges in eigenmaps to its continuous counterpart, the Laplace-Beltrami operator. Hence, a good sanity check is to compare the eigenmaps of the discrete and continuous Laplace operators. 

Firstly, we recall the eigenvalues of symmetric normalized graph Laplacians satisfies $0 \leq \lambda_k \leq 2$ for all $k$. As the eigenvalues can be interpreted as frequency components, the eigenvector associated to an high eigenvalue would correspond to an high frequency signal on the graph. To finish with, we propose a detailed analysis of the eigenmaps of our spaces of interest in terms of the Fourier basis and the spherical harmonics.

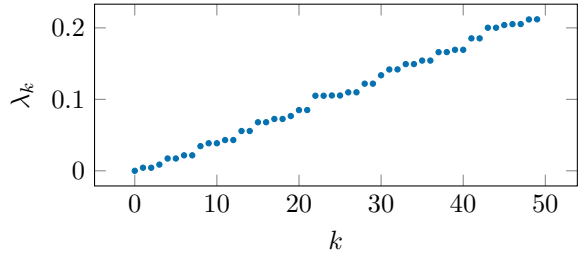
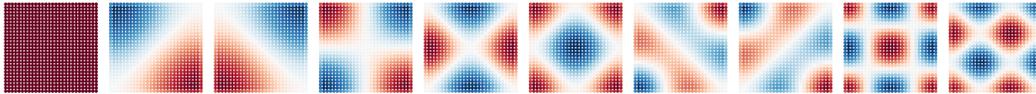
\begin{figure*}[h!]
    \centering
    \begin{subfigure}[b]{0.9\textwidth}
        \centering
        \begin{tikzpicture}
        \begin{axis}[
            width=8cm,
            height=4cm,
            xlabel={$k$},
            ylabel={$\lambda_k$},
            ]
            \addplot[only marks, color=color2, mark size=1] table {Data/r2_eigenval.dat};
        \end{axis}
        \end{tikzpicture}
        \caption{Eigenvalues of the $[0,1]^2$ space, from $\lambda_0$ to $\lambda_{49}$.}
    \end{subfigure}
    \hfill
    \begin{subfigure}[b]{\textwidth}
        \centering
        \includegraphics[width=\textwidth]{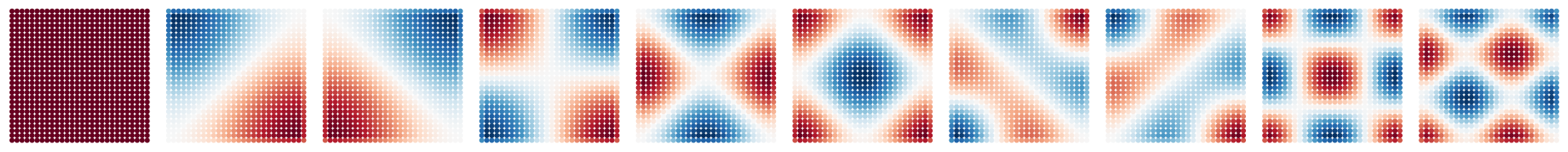}
        \caption{Eigenvectors of the $[0,1]^2$ space, from $\phi_0$ (left) to $\phi_9$ (right).}
    \end{subfigure}
    \caption{Eigenmaps of the $[0,1]^2$ space.}
    \label{fig:r2_eigenmaps}
\end{figure*}

\paragraph{Isotropic 2-dimensional grid.} The eigenmaps of the $[0,1)^2$ space are shown in figure \ref{fig:r2_eigenmaps}. It is a well-known fact that the 2-dimensional grid with periodic boundaries can be spanned by the Fourier basis. Consequently, the eigenvectors of this space corresponds to the sine and cosine trigonometric functions. In this case, the eigenvalues have multiplicity two (excepted the first one associated with a constant eigenvector), one for the sine and the other for the cosine. When the periodic condition at the boundaries are relaxed, the space is not homogeneous anymore, and the symmetries of the space change a little bit. The $[0,1)^2$ space is such a space where periodic condition does not hold. Hence, the eigenmaps of this space are not equal to the Fourier basis, but very close.

\begin{figure*}[h!]
    \centering
    \begin{subfigure}[b]{0.9\textwidth}
        \centering
        \begin{tikzpicture}
        \begin{axis}[
            width=8cm,
            height=4cm,
            xlabel={$k$},
            ylabel={$\lambda_k$},
            yticklabel style={
                    /pgf/number format/fixed,
                    /pgf/number format/precision=5
            },
            scaled y ticks=false
            ]
            \addplot[only marks, color=color2, mark size=1] table {Data/se2_eigenval.dat};
        \end{axis}
        \end{tikzpicture}
        \caption{Eigenvalues of the $[0,1]^2 \times [-\pi/2, \pi/2)$ space, from $\lambda_0$ to $\lambda_{49}$.}
    \end{subfigure}
    \hfill
    \begin{subfigure}[b]{\textwidth}
        \centering
        \includegraphics[width=\textwidth]{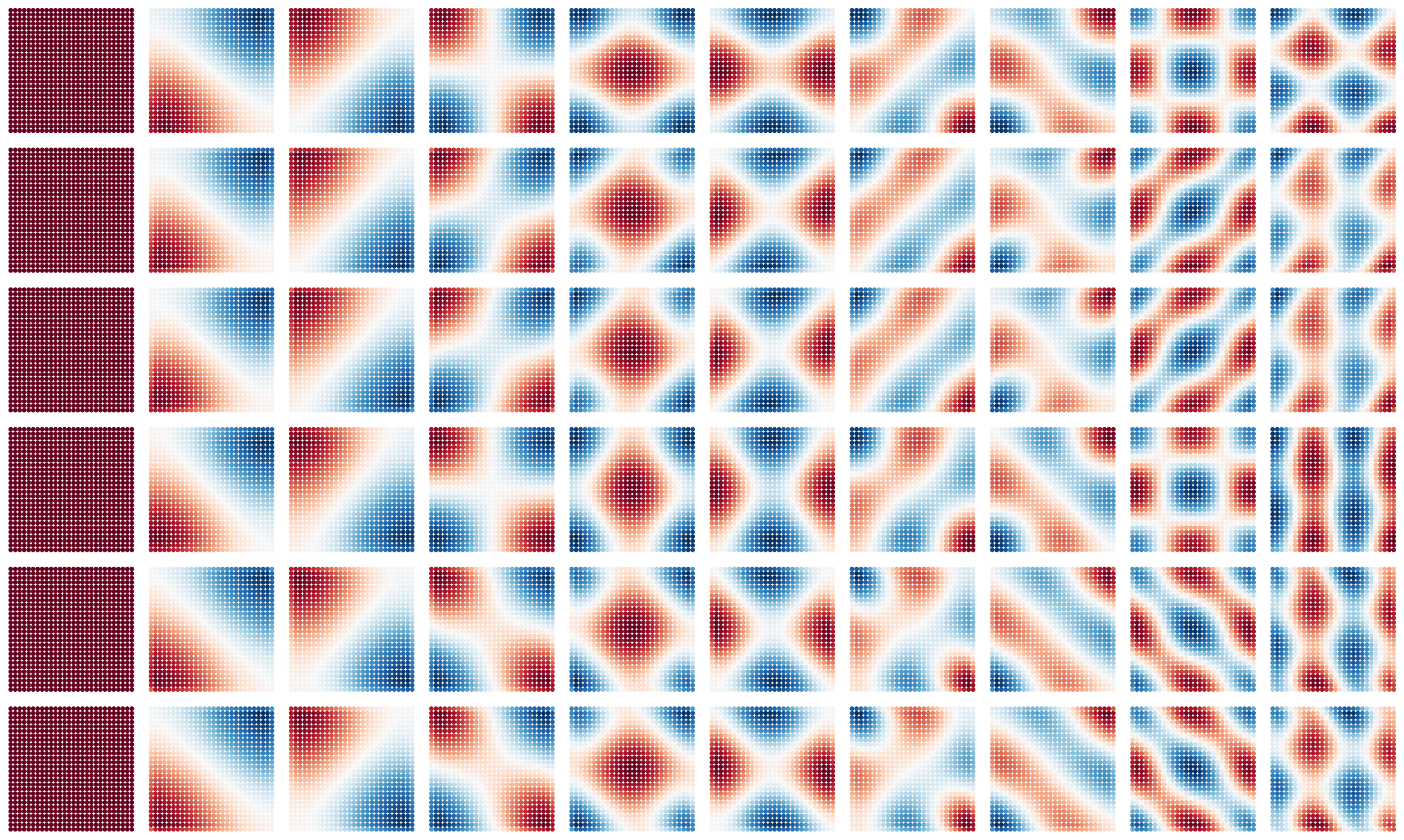}
        \caption{Eigenvectors of the $[0,1]^2 \times [-\pi/2, \pi/2)$ space, from $\phi_0$ (left) to $\phi_9$ (right).}
    \end{subfigure}
    \caption{Eigenmaps of the $[0,1]^2 \times [-\pi/2, \pi/2)$ space.}
    \label{fig:se2_eigenmaps}
\end{figure*}

\paragraph{Anisotropic 2-dimensional grid.} The eigenmaps of the $[0,1)^2 \times [-\pi/2, \pi/2)$ space are shown in figure \ref{fig:se2_eigenmaps}. We can extend the discussion we add in the isotropic case to justify the fact that the eigenmaps of the $[0,1)^2 \times [-\pi/2, \pi/2)$ is close to the Fourier basis. Moreover, we an additional orientation dimension, the rotation of the $[0,1)^2$ induces a torsion all along the orientation axis because of the anisotropic Riemannian metric.

\begin{figure*}[h!]
    \centering
    \begin{subfigure}[b]{0.9\textwidth}
        \centering
        \begin{tikzpicture}
        \begin{axis}[
            width=8cm,
            height=4cm,
            xlabel={$k$},
            ylabel={$\lambda_k$},
            ]
            \addplot[only marks, color=color2, mark size=1] table {Data/s2_eigenval.dat};
        \end{axis}
        \end{tikzpicture}
        \caption{Eigenvalues of the $S^2$ space, from $\lambda_0$ to $\lambda_{49}$.}
    \end{subfigure}
    \hfill
    \begin{subfigure}[b]{\textwidth}
        \centering
        \includegraphics[width=\textwidth]{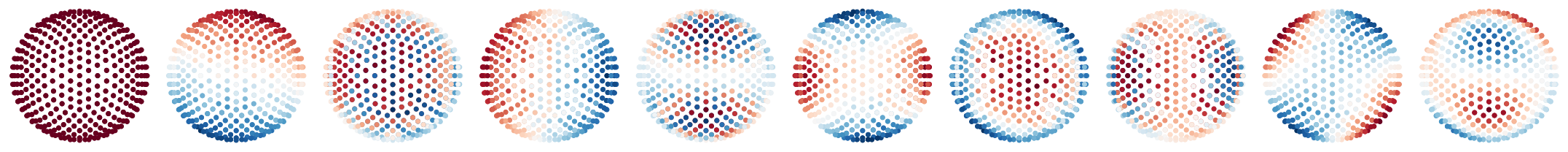}
        \caption{Eigenvectors of the $S^2$ space, from $\phi_0$ (left) to $\phi_9$ (right).}
    \end{subfigure}
    \caption{Eigenmaps of the $S^2$ space.}
    \label{fig:s2_eigenmaps}
\end{figure*}
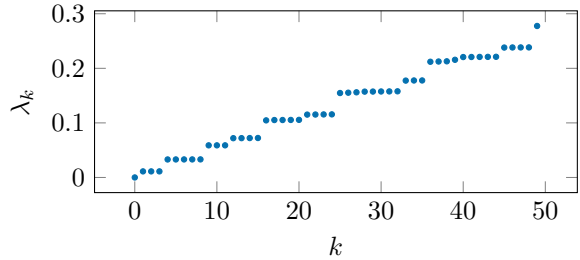
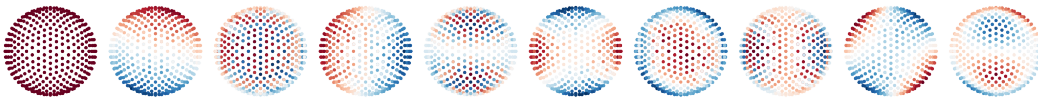

\paragraph{Isotropic 2-dimensional sphere.} The eigenmaps of the $S^2$ space are shown in figure \ref{fig:s2_eigenmaps}. It is a well-known fact that the 2-dimensional sphere is spanned by the spherical harmonics. The eigenvalues associated to this eigenvectors have increasing multiplicities of the form $2m + 1$ where $m$ is the order of the spherical harmonic.

\begin{figure*}[h!]
    \centering
    \begin{subfigure}[b]{0.9\textwidth}
        \centering
        \begin{tikzpicture}
        \begin{axis}[
            width=8cm,
            height=4cm,
            xlabel={$k$},
            ylabel={$\lambda_k$},
            ]
            \addplot[only marks, color=color2, mark size=1] table {Data/so3_eigenval.dat};
        \end{axis}
        \end{tikzpicture}
        \caption{Eigenvalues of the $S^2 \times [-\pi/2, \pi/2)$ space, from $\lambda_0$ to $\lambda_{49}$.}
    \end{subfigure}
    \hfill
    \begin{subfigure}[b]{\textwidth}
        \centering
        \includegraphics[width=\textwidth]{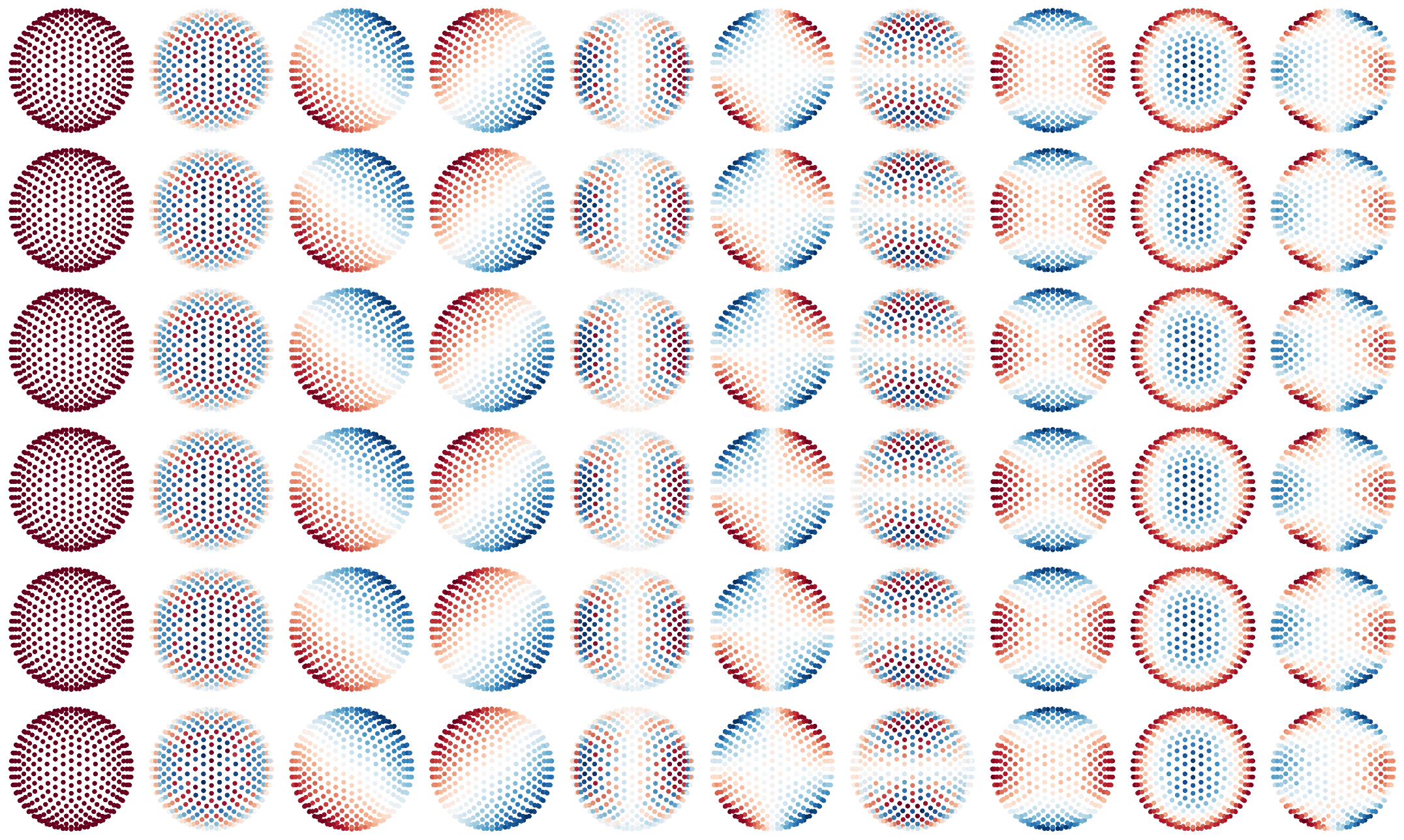}
        \caption{Eigenvectors of the $S^2 \times [-\pi/2, \pi/2)$ space, from $\phi_0$ (left) to $\phi_9$ (right).}
    \end{subfigure}
    \caption{Eigenmaps of the $S^2 \times [-\pi/2, \pi/2)$ space.}
    \label{fig:so3_eigenmaps}
\end{figure*}

\paragraph{Anisotropic 2-dimensional sphere.} The eigenmaps of the $S^2 \times [-\pi/2, \pi/2)$ space are shown in figure \ref{fig:so3_eigenmaps}. We can extend the discussion we add in the isotropic case to justify the fact that the eigenmaps of the $S^2 \times [-\pi/2, \pi/2)$ are close to the spherical harmonics. Nevertheless, the role of the orientation dimension is not easily interpretable. Indeed the notion of orientation is arduous to understand, as the kernel's orientation depends on the path, and not on this orientation location.

\end{document}